\documentclass{article}

\usepackage[T1]{fontenc}

\PassOptionsToPackage{table}{xcolor}
\usepackage{iclr2027_conference,times}

\usepackage{helvet}
\usepackage{tgtermes}
\usepackage[scale=1]{tgheros}
\usepackage{courier}
\usepackage[hyphens]{url}
\usepackage{graphicx}
\usepackage{caption}
\usepackage{subcaption}
\usepackage{amsmath,amssymb,mathtools}
\usepackage{booktabs}
\usepackage{xcolor}
\usepackage{algorithm}
\usepackage{algorithmic}
\usepackage{float}
\usepackage{wrapfig}

\usepackage[skins,breakable]{tcolorbox}

\usepackage[hidelinks]{hyperref}

\makeatletter
\let\JPEGDLM@orig@linkstart\hyper@linkstart
\def\hyper@linkstart#1#2{%
  \def\JPEGDLM@linktype{#1}\def\JPEGDLM@citetype{cite}%
  \ifx\JPEGDLM@linktype\JPEGDLM@citetype
    \def\@pdfborder{0 0 1}%
  \fi
  \JPEGDLM@orig@linkstart{#1}{#2}%
  \def\@pdfborder{0 0 0}%
}
\makeatother

\newcommand{\JPEGDLM}{\textsc{JPEG-DLM}}
\newcommand{\JPEGDLMFullName}{Joint-embedding Prediction for Efficient Generation with Diffusion Language Model}

\providecommand{\nolinenumbers}{}
\providecommand{\linenumbers}{}

\newcommand{\sg}{\mathrm{sg}}
\newcommand{\vectorsym}[1]{\boldsymbol{#1}}
\newcommand{\matrixsym}[1]{\boldsymbol{#1}}

\newcommand{\Ltok}{L}
\newcommand{\Llat}{N}

\newtcolorbox{generationexample}[1]{
  enhanced,
  breakable,
  colframe=gray!55,
  boxrule=0.55pt,
  arc=2mm,
  outer arc=2mm,
  interior style={left color=gray!3,right color=gray!12},
  left=10pt,
  right=10pt,
  top=9pt,
  bottom=9pt,
  before skip=8pt,
  after skip=8pt,
  fontupper=\ttfamily\footnotesize,
  before upper={%
    {\normalfont\sffamily\bfseries Example~#1\hfill
     \rmfamily\mdseries OWT-1024\par}
    \vspace{5pt}
    \raggedright
  }
}

\iclrfinalcopy

\title{\fontsize{15pt}{18pt}\selectfont
  One Latent, Many Tokens: Jointly Learning\\
  Compressed Embeddings for Efficient Language Diffusion}

\author{%
  \normalfont\small
  \begin{tabular*}{0.97\textwidth}{@{\extracolsep{\fill}}ccc@{}}
    \parbox[t]{0.27\textwidth}{\centering
      \textbf{Yulin Yuan} \\
      Zhejiang University \\
      yulinyuan@zju.edu.cn
    }
    &
    \parbox[t]{0.28\textwidth}{\centering
      \textbf{Ying Zhang} \\
      University of Cambridge \\
      yz2160@cam.ac.uk
    }
    &
    \parbox[t]{0.34\textwidth}{\centering
      \textbf{Xiangming Meng\thanks{Corresponding author.}} \\
      \resizebox{\linewidth}{!}{ZJU-UIUC Institute, Zhejiang University} \\
      xiangmingmeng@intl.zju.edu.cn
    }
  \end{tabular*}
}

\hypersetup{
  pdftitle={One Latent, Many Tokens: Jointly Learning Compressed Embeddings for Efficient Language Diffusion},
  pdfauthor={Yulin Yuan, Ying Zhang, Xiangming Meng}
}

\begin{document}
\maketitle
\lhead{Preprint}

\raggedbottom

\begin{abstract}
Most continuous diffusion language models process one latent position per token at each sampling step, making generation expensive.
Two-stage methods lower the cost by reducing the latent length, but they fix the compressed embedding space before training the diffusion model.
Embeddings from the fixed space can be difficult to model with diffusion and decode reliably into tokens, which limits generation quality after compression.
To address this problem, we introduce \JPEGDLM{} (\JPEGDLMFullName{}), which jointly trains a compressor, a flow matching model and a decoding module.
With joint-embedding prediction, \JPEGDLM{} learns compressed embeddings that are more structured, easier to model with diffusion and reliably decodable into tokens.
\JPEGDLM{} achieves the highest throughput among recent diffusion and flow models on LM1B and OWT, with a Gen-PPL lower than most of them.
At a compression rate of $0.5$ on OWT, it reaches a Gen-PPL of $34.52$ and approximately $2.3\times$ ELF's throughput.
These results suggest that jointly learning compressed embeddings offers a promising path toward efficient diffusion language modeling.
Code is available at \url{https://github.com/infiniteYuanyl/JPEG-DLM}.
\end{abstract}

\suppressfloats[t]
\begin{figure}[t]
\captionsetup{aboveskip=2pt,belowskip=0pt}
\centering
\includegraphics[width=\linewidth]{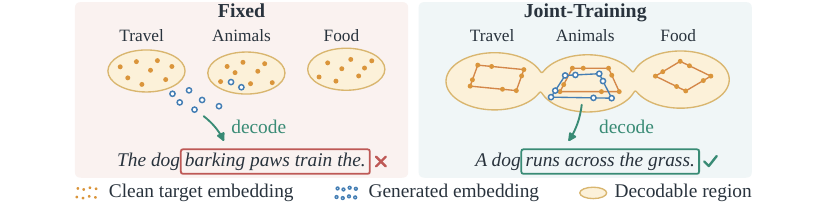}
\caption{Schematic comparison of fixed and jointly learned compressed spaces, showing more structured embeddings and broader decodable regions with joint training.}
\label{fig:insight-banner}
\end{figure}

\section{Introduction}

Diffusion language models (DLMs) generate text by iteratively refining a corrupted sequence and can update multiple token positions in parallel~\citep{nie2025llada,ye2025dream}. Existing methods operate either in discrete token space~\citep{austin2021d3pm,lou2024sedd} or continuous representation space~\citep{li2022diffusionlm,hu2026elf}. Masked diffusion, a prominent discrete approach, has achieved competitive language modeling performance, including at large-model scale~\citep{sahoo2024mdlm,shi2024md4,nie2025llada}. Continuous DLMs instead encode tokens as continuous features, connect the data and noise distributions through a corruption process or conditional probability path~\citep{li2022diffusionlm,lipman2023flowmatching}, and decode the generated representation back into tokens~\citep{li2022diffusionlm,han2023ssdlm,hu2026elf,lovelace2023ld4lg}. Continuous states support direct gradient-based guidance~\citep{li2022diffusionlm}, while learned latents can aggregate context across token positions~\citep{guo2026cola}. The same flexibility underlies latent image diffusion~\citep{rombach2022latentdiffusion} and joint text--image diffusion~\citep{bao2023unidiffuser}.

Parallel updates do not, by themselves, make generation inexpensive. Autoregressive language models generate one token per step and are inherently sequential~\citep{radford2019language}, whereas discrete DLMs can update multiple positions~\citep{sahoo2024mdlm,nie2025llada}. Most continuous DLMs, however, assign one latent position to each token~\citep{li2022diffusionlm,hu2026elf,chen2026langflow}. Because self-attention computation grows quadratically with sequence length, reducing latent length provides quadratic savings in attention computation at each model evaluation. However, stronger compression can make it harder to preserve generation quality. This raises the question of how to reduce the latent length used for generation without substantially compromising generation quality.

Prior work~\citep{meshchaninov2025cosmos,liang2026auroralm} on latent-length compression for diffusion language models uses two-stage training. It first trains an encoder and decoder to reconstruct text from compressed embeddings and then trains a generative model on the fixed embedding distribution. A fixed embedding space can limit how well the generative model learns the text distribution. Recent work~\citep{meshchaninov2026ldlm} instead jointly trains the latent encoder, diffusion model, and decoder, so the generative objective also updates the representation. A natural idea, as illustrated in Figure~\ref{fig:insight-banner}, is therefore to learn compressed embeddings jointly with the flow matching model so that they support generation and remain reliably decodable into tokens.

Inspired by this insight, we introduce \JPEGDLM{}, which jointly trains a compressor, a flow matching model, and a decoding module for efficient text generation with diffusion language models. In particular, a fixed compressed embedding space does not ensure that clean embeddings can be recovered from corrupted compressed views. Drawing on JEPA~\citep{lecun2022path,assran2023ijepa}, which predicts one embedding from another, joint-embedding prediction (JEP) trains the context compressor and shared-weight network to recover the clean target embedding from a corrupted compressed view. In the denoise branch, flow matching~\citep{lipman2023flowmatching} trains the shared-weight network to recover clean embeddings from noisy embeddings formed by interpolating clean target embeddings with Gaussian noise. The clean-embedding prediction from the decode branch then enters the decoding module. Feature reconstruction and token-level cross-entropy train this module to map the predicted clean embedding to tokens reliably. Compared with methods that use a fixed compressed embedding space, \JPEGDLM{} jointly learns the compressed embeddings with the generative model and achieves a marked improvement in text generation quality. Experiments on LM1B and OWT further show that it achieves the highest throughput among diffusion and flow models, with a Gen-PPL lower than most of them.

The main contributions of this paper are as follows:
\begin{itemize}
  \item We introduce \JPEGDLM{}, which learns latent-length compression jointly with generation. This yields more structured compressed embeddings that are easier to model with diffusion and decode into tokens reliably.
  \item We use JEP to predict clean target embeddings from corrupted compressed views, training the compressor to preserve structure that remains predictable under corruption.
  \item Under latent-length compression, \JPEGDLM{} reaches generation quality comparable to the baselines on LM1B and OWT while achieving the highest generation throughput.
\end{itemize}

\section{Related Work}

\subsection{Diffusion and Flow Language Models}
\label{sec:rw-dlm}

Diffusion language models generate text by iteratively transforming a corrupted state into a clean sample. Discrete methods define this process directly over token categories, using either general transition matrices~\citep{austin2021d3pm,lou2024sedd} or an absorbing mask state~\citep{shi2024md4,sahoo2024mdlm}. In masked diffusion, a forward process replaces tokens with a mask symbol, while generation uses learned reverse transition probabilities to recover masked positions iteratively. The generative state therefore remains categorical at every token position throughout sampling. Continuous methods instead map tokens into continuous representations and denoise in the resulting space, using token embeddings in Diffusion-LM~\citep{li2022diffusionlm} and Plaid~\citep{gulrajani2023plaid}, or a simplex defined over the vocabulary in SSD-LM~\citep{han2023ssdlm} and TESS~\citep{mahabadi2024tess}. Flow matching~\citep{lipman2023flowmatching} defines a probability path between the noise and data distributions, and learns the velocity field of this path at every continuous time $t$. An ODE solver then integrates this field to transport noise into the embedding distribution. Recent flow models learn their own token embeddings in LangFlow~\citep{chen2026langflow}, or reuse contextual features from a frozen encoder in ELF~\citep{hu2026elf}, and reach generation quality that rivals discrete diffusion. Like these flow models, \JPEGDLM{} uses flow matching to generate continuous embeddings, which are then decoded into tokens.

\subsection{Efficient Generation in Diffusion Language Models}
\label{sec:rw-eff}

A single forward pass of a discrete diffusion model predicts all masked positions at once, so a sampler can unmask the positions it is most confident about and leave the rest masked for later steps. Decoding methods for these models select which positions to commit at each step~\citep{yuan2026vrcd,qi2026clad}, and committing several positions per pass raises throughput while keeping task accuracy close to the backbone~\citep{wu2026fastdllm}. Continuous models decode only after the final state is reached~\citep{hu2026elf,meshchaninov2026ldlm}, so no position is fixed while the sequence is still being refined, and this form of parallel commitment is not suited to them. Their generation cost instead follows the number of latent positions processed at each step, so a shorter latent sequence offers a means of reducing it~\citep{lovelace2023ld4lg,liang2026auroralm}.

LD4LG~\citep{lovelace2023ld4lg} compresses frozen encoder features into a fixed-length latent sequence with a Perceiver resampler, and PLANNER~\citep{zhang2023planner} encodes a paragraph into 16 latent codes with a variational encoder. Both decode text with a pretrained autoregressive decoder, so generation depends on the preceding tokens as well as the latent representation. COSMOS~\citep{meshchaninov2025cosmos} removes this dependence by training a Perceiver-resampler autoencoder whose decompressor and token projection decode the sequence directly. AURORA-LM~\citep{liang2026auroralm} builds an ordered latent sequence with a causal autoencoder and sets its latent length as a fixed fraction of the token length. In all of these methods the embedding space is fixed before the diffusion model is trained, and such a space may limit the generation quality that can be reached. \JPEGDLM{} instead learns the compressed embedding space jointly with the flow matching model and the decoding module to reach better generation quality.

\subsection{Learning an Embedding Space with Joint-Embedding Prediction}
\label{sec:rw-jepa}

Latent diffusion language models depend on the quality of their embedding space~\citep{meshchaninov2026ldlm}. Joint-embedding predictive architectures~\citep{lecun2022path} approach this as a representation learning problem. They first corrupt an input by masking part of it or adding noise. A context encoder maps the corrupted view to a context embedding. A predictor then uses this context embedding to predict the embedding produced by a target encoder for the clean input~\citep{assran2023ijepa,baevski2022data2vec}. The loss is thus computed in embedding space, not input space. The target encoder is an exponential moving average of the context encoder, and its output enters the loss with a stop-gradient. Without this asymmetry the objective admits a trivial solution in which the encoder produces a constant representation regardless of its input~\citep{bardes2024vjepa}. These architectures have been instantiated across modalities, including speech, vision and language in data2vec~\citep{baevski2022data2vec}, video in V-JEPA~\citep{bardes2024vjepa}, image generation in D-JEPA~\citep{chen2025djepa}, and text in LLM-JEPA~\citep{huang2026llmjepa}.

\JPEGDLM{} uses joint-embedding prediction to learn a compressed embedding space and trains a flow matching model to generate from it. Recent methods, such as LDLM~\citep{meshchaninov2026ldlm} and Cola DLM~\citep{guo2026cola}, also train the latent space together with its generative model. In both, the generative objective reaches the encoder at every noise level, and each adds a regularizer to keep the representation from drifting. \JPEGDLM{} updates the compressor through the $t=1$ decode branch, while the denoise branch trains the shared-weight network using stop-gradient compressed embeddings as targets. At high noise levels, the predicted clean embedding approaches the data mean and carries little information about individual tokens~\citep{karras2022edm}. Token-level cross-entropy on such a prediction would drive large gradients through the token projection and destabilize training.

\section{Method}


\newcommand{\JPEGAlgorithmComment}[1]{{\color[rgb]{0.18,0.49,0.20}\texttt{\#}\enspace#1}}
\newcommand{\JPEGAlgorithmFeatureMatrix}{\matrixsym{h}}
\newcommand{\JPEGAlgorithmPredictedFeatureMatrix}{\widehat{\matrixsym{h}}}
\newcommand{\JPEGInferenceSectionGap}{\STATE \mbox{}\vspace{1.059pt}}

\newcommand{\JPEGDLMTrainingProcedure}{%
\begin{minipage}[t]{\linewidth}
\captionsetup{format=plain,labelfont=bf,labelsep=space,
  justification=raggedright,singlelinecheck=false,
  aboveskip=0pt,belowskip=0.15em}
\hrule height 0.6pt
\captionof{algorithm}{\JPEGDLM{}: training.}
\label{alg:training}
\hrule height 0.4pt
\vspace{0.4em}
{\footnotesize\ttfamily
\begin{algorithmic}
\STATE \JPEGAlgorithmComment{Input: $\vectorsym{x}$; frozen $E$; $\sigma,\pi$}
\STATE \JPEGAlgorithmComment{Models: $C_\theta,\bar C,f_\phi,D_\psi,W$}
\STATE $\JPEGAlgorithmFeatureMatrix\leftarrow E(\vectorsym{x})$
\STATE $\vectorsym{z}_{\mathrm{tgt}}\leftarrow\sg\big(\bar C(\JPEGAlgorithmFeatureMatrix)\big)$
\STATE \mbox{}
\STATE \JPEGAlgorithmComment{decode branch ($t=1$)}
\STATE $\vectorsym{\epsilon}_h\sim\mathcal{N}(0,\matrixsym{I})$
\STATE $\vectorsym{z}_c\leftarrow C_\theta(\JPEGAlgorithmFeatureMatrix+\sigma\cdot\vectorsym{\epsilon}_h)$
\STATE $\widehat{\vectorsym{z}}_{\mathrm{tgt}}\leftarrow f_\phi(\vectorsym{z}_c,t{=}1)$
\STATE $\mathcal{L}_{\mathrm{jep}}\leftarrow\|\widehat{\vectorsym{z}}_{\mathrm{tgt}}-\vectorsym{z}_{\mathrm{tgt}}\|_2^2$
\STATE $\JPEGAlgorithmPredictedFeatureMatrix\leftarrow D_\psi(\widehat{\vectorsym{z}}_{\mathrm{tgt}})$
\STATE $\mathcal{L}_{\mathrm{feat}}\leftarrow\|\JPEGAlgorithmPredictedFeatureMatrix-\sg(\JPEGAlgorithmFeatureMatrix)\|_2^2$
\STATE $\vectorsym{\ell}\leftarrow W(\JPEGAlgorithmPredictedFeatureMatrix)$
\STATE $\mathcal{L}_{\mathrm{ce}}\leftarrow\operatorname{CE}(\vectorsym{\ell},\vectorsym{x})$
\STATE \mbox{}
\STATE \JPEGAlgorithmComment{denoise branch}
\STATE $t\sim\pi$; $\vectorsym{\epsilon}\sim\mathcal{N}(0,\matrixsym{I})$
\STATE $\vectorsym{z}_t\leftarrow t\vectorsym{z}_{\mathrm{tgt}}+(1-t)\vectorsym{\epsilon}$
\STATE $\mathcal{L}_{\mathrm{flow}}\leftarrow\|f_\phi(\vectorsym{z}_t,t)-\vectorsym{z}_{\mathrm{tgt}}\|_2^2$
\end{algorithmic}
}
\hrule height 0.6pt
\end{minipage}
}

\newcommand{\JPEGDLMInferenceProcedure}{%
\begin{minipage}[t]{\linewidth}
\captionsetup{format=plain,labelfont=bf,labelsep=space,
  justification=raggedright,singlelinecheck=false,
  aboveskip=0pt,belowskip=0.15em}
\hrule height 0.6pt
\captionof{algorithm}{\JPEGDLM{}: inference.}
\label{alg:inference}
\hrule height 0.4pt
\vspace{0.4em}
{\footnotesize\ttfamily
\begin{algorithmic}
\STATE \JPEGAlgorithmComment{Models: trained $f_\phi,D_\psi,W$}
\STATE \JPEGAlgorithmComment{State $(\Llat,d)$; times $t_0=0,\ldots,t_K=1$}
\STATE $\vectorsym{z}\sim\mathcal{N}(0,\matrixsym{I})$
\JPEGInferenceSectionGap
\STATE \mbox{}
\STATE \JPEGAlgorithmComment{start sampling}
\FOR{$k=0$ \TO $K-1$}
  \STATE $\widehat{\vectorsym{z}}_1\leftarrow f_\phi(\vectorsym{z},t{=}t_k)$
  \STATE $\vectorsym{v}\leftarrow(\widehat{\vectorsym{z}}_1-\vectorsym{z})/(1-t_k)$
  \STATE $\vectorsym{z}\leftarrow\vectorsym{z}+(t_{k+1}-t_k)\vectorsym{v}$
\ENDFOR
\JPEGInferenceSectionGap
\STATE \JPEGAlgorithmComment{predict clean embedding at $t=1$}
\STATE $\widehat{\vectorsym{z}}_{\mathrm{dec}}\leftarrow f_\phi(\vectorsym{z},t{=}1)$
\JPEGInferenceSectionGap
\STATE \JPEGAlgorithmComment{decode tokens}
\STATE $\JPEGAlgorithmPredictedFeatureMatrix\leftarrow D_\psi(\widehat{\vectorsym{z}}_{\mathrm{dec}})$
\STATE $\vectorsym{\ell}\leftarrow W(\JPEGAlgorithmPredictedFeatureMatrix)$
\STATE $\widehat{x}_i\leftarrow\operatorname*{arg\,max}_{v\in\mathcal{V}}\ell_{i,v},\quad i=1,\ldots,\Ltok$
\end{algorithmic}
}
\hrule height 0.6pt
\end{minipage}
}

\newcommand{\JPEGDLMProcedures}{%
\begin{figure}[t]
\begin{minipage}[t]{0.49\textwidth}
\JPEGDLMTrainingProcedure
\end{minipage}\hfill
\begin{minipage}[t]{0.49\textwidth}
\JPEGDLMInferenceProcedure
\end{minipage}
\end{figure}
}

\begin{figure}[t]
\centering
\includegraphics[width=\textwidth]{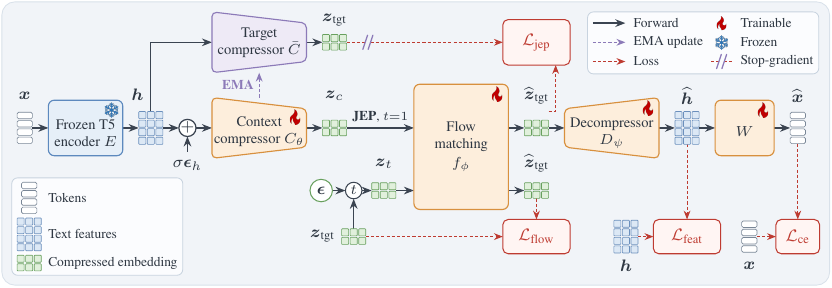}
\caption{Overview of \JPEGDLM{}. The frozen T5 encoder maps tokens to features. The context compressor maps a corrupted view to $\vectorsym{z}_c$, while the target compressor provides the stop-gradient clean target embedding $\vectorsym{z}_{\mathrm{tgt}}$ and is updated by an exponential moving average. The shared-weight network predicts the clean target embedding from $\vectorsym{z}_c$ for JEP and from the noisy embedding $\vectorsym{z}_t$ for flow matching. The decoding module maps the predicted clean embedding back to features and tokens using $\mathcal{L}_{\mathrm{feat}}$ and $\mathcal{L}_{\mathrm{ce}}$.}
\label{fig:arch}
\end{figure}

We introduce \JPEGDLM{}, a joint training method for efficient text generation in a continuous latent space with reduced latent length. As illustrated in Figure~\ref{fig:arch}, \JPEGDLM{} uses joint-embedding prediction to shape the compressed latent space, applies flow matching to generate compressed embeddings, and trains a decoding module to map predicted clean embeddings to tokens. Section~\ref{sec:method-preliminaries} first introduces the background of latent diffusion and flow matching. Section~\ref{sec:method-jep} then explains how joint-embedding prediction shapes the compressed latent space under latent-length compression. Section~\ref{sec:method-decode} presents joint training for embedding generation and reliable token decoding.

\subsection{Preliminaries}
\label{sec:method-preliminaries}

Latent diffusion models move iterative generation from the original data space to a continuous latent representation~\citep{rombach2022latentdiffusion}. For text generation, an encoder maps a token sequence $\vectorsym{x}$ to a continuous latent $\vectorsym{z}_1$, a latent generative model learns the distribution of $\vectorsym{z}_1$, and a decoding module maps generated latents back to tokens~\citep{lovelace2023ld4lg}. We write $\vectorsym{z}_1\in\mathbb{R}^{\Llat\times d}$, where $\Llat$ denotes latent length and $d$ denotes latent width. When $\Llat<\Ltok$, the generative model processes fewer latent positions than the original token sequence at each sampling step.

We use flow matching~\citep{lipman2023flowmatching} to learn the latent distribution by transporting a simple noise distribution $p_0$ to the latent data distribution $p_1$ along a continuous probability path. Given a clean embedding $\vectorsym{z}_1\sim p_1$ and Gaussian noise $\vectorsym{\epsilon}\sim\mathcal{N}(0,\matrixsym{I})$, we construct the noisy embedding $\vectorsym{z}_t$ by linear interpolation,
\begin{equation}
\label{eq:flow-path}
\vectorsym{z}_t=t\vectorsym{z}_1+(1-t)\vectorsym{\epsilon}, \qquad t\in[0,1],
\end{equation}
Here, $t=0$ corresponds to Gaussian noise and $t=1$ corresponds to the clean embedding. Following JiT~\citep{li2026jit}, we use $x$-prediction, where the flow matching model $f_\phi$ predicts the clean embedding $\vectorsym{z}_1$ from the noisy embedding $\vectorsym{z}_t$ instead of directly regressing the velocity. With $t\sim\pi$, where $\pi$ is a logit-normal distribution over $t$, we train this embedding prediction using the mean squared error $\|\cdot\|_2^2$ over positions and channels
\begin{equation}
\label{eq:flow-loss}
\mathcal{L}_{\mathrm{flow}}
=\mathbb{E}_{\substack{\vectorsym{z}_1\sim p_1,\;t\sim\pi,\\
\vectorsym{\epsilon}\sim\mathcal{N}(0,\matrixsym{I})}}
\left[\left\|f_\phi(\vectorsym{z}_t,t)-\vectorsym{z}_1\right\|_2^2\right],
\end{equation}
At inference, an Euler ODE solver~\citep{li2026jit} performs $K$ evaluations of $f_\phi$ to evolve Gaussian noise to the terminal state. We then evaluate $f_\phi$ once at $t=1$ to obtain the clean embedding for decoding, which the decoding module maps back to the token sequence $\vectorsym{x}$. A detailed derivation of the induced velocity field is provided in Appendix~\ref{app:clean-target-velocity}, and the sampling FLOPs under latent-length compression are reported in Table~\ref{tab:app-system-config}.

\JPEGDLMProcedures

\subsection{Joint-Embedding Prediction with Latent-Length Compression}
\label{sec:method-jep}

Our first goal is to learn a compressed latent space that reduces latent length, preserves the information needed for reliable token decoding, and allows clean embeddings to be predicted under corruption. A frozen pretrained T5 encoder~\citep{raffel2020t5} maps the token sequence to the feature matrix $\matrixsym{h}=E(\vectorsym{x})\in\mathbb{R}^{\Ltok\times d}$. We use a Perceiver-style compressor~\citep{alayrac2022flamingo} that maps its $\Ltok$ token positions to $\Llat$ latent positions. Specifically, $\Llat$ learned latent queries cross-attend to all $\Ltok$ token-level features, and the resulting query states form the compressed embedding $\vectorsym{z}\in\mathbb{R}^{\Llat\times d}$. We define the latent-length compression rate as $r=\Llat/\Ltok$, where $r=1.0$ preserves one latent position per token and a smaller value indicates stronger compression. To restore token-length features, the decompressor $D_\psi$ uses $\Ltok$ token-position queries to cross-attend to all $\Llat$ compressed positions, allowing each restored position to use information from the complete compressed embedding. Because both mappings attend over their complete inputs, the compressor and decompressor process the sequence bidirectionally.

However, the architecture of the compressor and decompressor does not by itself learn a latent space in which clean embeddings can be predicted under corruption. Following JEPA~\citep{lecun2022path,assran2023ijepa}, we train a predictor to recover a clean compressed embedding from a corrupted compressed view. Specifically, we retain $\matrixsym{h}$ as the clean feature matrix and form the corrupted view $\widetilde{\matrixsym{h}}=\matrixsym{h}+\sigma\vectorsym{\epsilon}_h$, where $\vectorsym{\epsilon}_h\sim\mathcal{N}(0,\matrixsym{I})$, after which the context compressor maps $\widetilde{\matrixsym{h}}$ to the compressed embedding $\vectorsym{z}_c=C_\theta(\widetilde{\matrixsym{h}})$. The predictor estimates the corresponding clean compressed embedding $\widehat{\vectorsym{z}}_{\mathrm{tgt}}$ from $\vectorsym{z}_c$, while its stop-gradient target $\sg(\bar C(\matrixsym{h}))$ is computed by a copy $\bar C$ whose parameters are updated as an exponential moving average of $C_\theta$ to prevent representation collapse~\citep{assran2023ijepa,bardes2024vjepa}. We define the JEP loss as the mean squared error between the predicted embedding and its stop-gradient target.
\begin{equation}
\label{eq:jep}
\mathcal{L}_{\mathrm{jep}}
=\left\|\widehat{\vectorsym{z}}_{\mathrm{tgt}}-\sg(\bar C(\matrixsym{h}))\right\|_2^2.
\end{equation}
This loss updates the predictor and context compressor, encouraging $\vectorsym{z}_c$ to preserve the information needed to recover the clean target embedding under corruption. Overall, we use joint-embedding prediction to learn a compressed latent space whose representations have reduced latent length and allow clean embeddings to be predicted under corruption.

\subsection{Joint Training for Embedding Generation and Reliable Token Decoding}
\label{sec:method-decode}

We use the flow matching model $f_\phi$ in the denoise branch to model the distribution of the compressed embeddings learned in Section~\ref{sec:method-jep}. The denoise branch samples $t\sim\pi$, constructs the noisy embedding $\vectorsym{z}_t$ using the interpolation in Equation~\ref{eq:flow-path}, and minimizes $\mathcal{L}_{\mathrm{flow}}$ in Equation~\ref{eq:flow-loss} to train $f_\phi$ to recover the clean target embedding $\vectorsym{z}_{\mathrm{tgt}}=\sg(\bar C(\matrixsym{h}))$.

To make the generated compressed embeddings easier to decode into tokens, we follow ELF~\citep{hu2026elf} and share the parameters of $f_\phi$ between the JEP predictor and the flow matching model with $x$-prediction. Model-mode tokens identify the denoise and decode branches, which use separate output heads. At $t=1$, the decode branch receives $\vectorsym{z}_c$, whose corruption is controlled by $\sigma$ independently of the flow time. It predicts the clean target embedding $\widehat{\vectorsym{z}}_{\mathrm{tgt}}=f_\phi(\vectorsym{z}_c,t{=}1)$, which is supervised by $\mathcal{L}_{\mathrm{jep}}$ in Equation~\ref{eq:jep}. The decoding module then receives $\widehat{\vectorsym{z}}_{\mathrm{tgt}}$ and produces token logits $\vectorsym{\ell}$. We apply token-level cross-entropy to these logits only in the decode branch at $t=1$ and train the decoding module with
\begin{equation}
\label{eq:ce}
\mathcal{L}_{\mathrm{ce}}
=-\frac{1}{\Ltok}\sum_{i=1}^{\Ltok}
\log\frac{\exp(\ell_{i,x_i})}{\sum_{v\in\mathcal{V}}\exp(\ell_{i,v})}.
\end{equation}
Here, $\mathcal{V}$ denotes the token vocabulary. Once ODE sampling reaches the final state, we evaluate $f_\phi(\cdot,1)$ once in the decode branch, and the decoding module receives the predicted clean embedding. This gives the decoding module the same type of input during training and inference.

We first use the decompressor $D_\psi$ to restore the token-length feature matrix $\widehat{\matrixsym{h}}=D_\psi(\widehat{\vectorsym{z}}_{\mathrm{tgt}})$. A token projection $W$ then maps these features to token logits $\vectorsym{\ell}=W(\widehat{\matrixsym{h}})$. The projection consists of two linear layers with a GELU activation between them. We supervise the decompressed features directly with a feature reconstruction loss before applying the token projection
\begin{equation}
\label{eq:feat}
\mathcal{L}_{\mathrm{feat}}
=\left\|\widehat{\matrixsym{h}}-\sg(\matrixsym{h})\right\|_2^2.
\end{equation}
By supervising $\widehat{\matrixsym{h}}$ before the token projection, $\mathcal{L}_{\mathrm{feat}}$ provides a direct training signal to the decompressor.

Accordingly, $\mathcal{L}_{\mathrm{flow}}$ trains the model to generate compressed embeddings, $\mathcal{L}_{\mathrm{jep}}$ shapes the compressed representation, $\mathcal{L}_{\mathrm{ce}}$ supervises token decoding, and $\mathcal{L}_{\mathrm{feat}}$ stabilizes feature reconstruction. We sum these four losses to obtain the main objective
\begin{equation}
\label{eq:main-loss}
\mathcal{L}_{\mathrm{main}}
=\mathcal{L}_{\mathrm{jep}}
+\mathcal{L}_{\mathrm{flow}}
+\mathcal{L}_{\mathrm{feat}}
+\mathcal{L}_{\mathrm{ce}}.
\end{equation}
We also use additional regularization losses described in Appendix~\ref{app:sigreg}.
The denoise branch updates $f_\phi$ through $\mathcal{L}_{\mathrm{flow}}$ using stop-gradient target embeddings. The decode branch updates $C_\theta$ and $f_\phi$ through $\mathcal{L}_{\mathrm{jep}}$, $\mathcal{L}_{\mathrm{feat}}$, and $\mathcal{L}_{\mathrm{ce}}$. The feature and token losses also update the decoding module. The context compressor therefore receives no gradients from the denoise branch, and flow matching affects it only through the shared $f_\phi$.

For further details of the training and inference procedures, see the pseudocode in Algorithms~\ref{alg:training} and~\ref{alg:inference}, respectively.

\section{Experiments}
\label{sec:experiments}

\noindent\textbf{Datasets.} We evaluate unconditional generation on the One Billion Word Benchmark (LM1B)~\citep{chelba2014one} and OpenWebText (OWT)~\citep{gokaslan2019openwebtext}, using sequence lengths of 128 and 1024, respectively. We also use OWT with a sequence length of 128 (OWT-128) to analyze compressed latent representations and conduct ablation studies.

\noindent\textbf{Evaluation.} We report generative perplexity (Gen-PPL), the perplexity of generated text under a pretrained GPT-2 Large~\citep{radford2019language}, and average unigram entropy as a measure of diversity. A low Gen-PPL can also come from degenerate text~\citep{franca2026hacking}, so we report MAUVE~\citep{pillutla2021mauve}, which compares the generated and reference distributions and reflects generation quality more faithfully. The main comparison also reports generation speedup relative to the AR baseline on each dataset.

\noindent\textbf{Models and baselines.} \JPEGDLM{} uses a frozen T5-small encoder~\citep{raffel2020t5}, a six-layer compressor, and a six-layer decompressor with latent width $d=512$. We use a DiT~\citep{peebles2023dit} as the flow matching model. The DiT and token projection have 105M parameters. We compare \JPEGDLM{} with an autoregressive Transformer (AR), the discrete diffusion models MDLM~\citep{sahoo2024mdlm}, SEDD-Absorb~\citep{lou2024sedd}, and Duo~\citep{sahoo2025duo}, the flow matching models FLM~\citep{lee2026flowmap}, LangFlow~\citep{chen2026langflow}, and ELF~\citep{hu2026elf}, the jointly trained latent diffusion model LDLM~\citep{meshchaninov2026ldlm}, and the two-stage latent diffusion model COSMOS~\citep{meshchaninov2025cosmos}.

\noindent\textbf{Training and inference.} We train \JPEGDLM{} in bfloat16 with the AdamW optimizer~\citep{loshchilov2019decoupled}. The flow matching model uses a learning rate of $1\times10^{-3}$ on both datasets. The compressor and decompressor both use $3\times10^{-4}$ on LM1B and $2\times10^{-4}$ on OWT. During inference, we use self-conditioning~\citep{chen2023analogbits} and the Euler ODE sampler for 32 steps on both datasets, followed by greedy token decoding. Appendix~\ref{app:experimental-details} lists the model configurations, training schedules, and evaluation details.

\subsection{Main Comparison}
\label{sec:exp-setup}
\label{sec:exp-main}

\JPEGDLM{} uses $r=0.25$ on LM1B and $r=0.5$ on OWT, reducing 128 and 1024 token positions to 32 and 512 latent positions, respectively. On LM1B, we also report $r=0.5$, which uses 64 latent positions. We report the mean over five seeds, with 1{,}024 samples per seed. MDLM, SEDD-Absorb, Duo, FLM, LDLM, and LangFlow use a common sampling budget of 128 steps on LM1B and 1024 steps on OWT. ELF~\citep{hu2026elf} reports its main unconditional result with 32 sampling steps. We retain this step count with its ODE sampler, and \JPEGDLM{} also uses 32-step ODE sampling. We also report ELF and \JPEGDLM{} with ELF's 32-step SDE sampler.

On both datasets, \JPEGDLM{} achieves  a Gen-PPL close to that of, and the highest throughput among, the diffusion and flow models in Table~\ref{tab:main-uncond}. We use AR without KV cache as the speed reference. At 32 ODE sampling steps, \JPEGDLM{} achieves lower Gen-PPL than ELF. It also uses  no more sampling steps than the remaining diffusion and flow models. On OWT, \JPEGDLM{} reaches $2.31\times$ the throughput of ELF and $4.69\times$ that of AR with KV cache.

\begin{table}[H]
\caption{Unconditional generation on LM1B and OWT. Speed is relative to AR without KV cache. Among diffusion and flow models, \textbf{bold} and \underline{underlined} mark the best and second-best Gen-PPL and speed. \JPEGDLM{} has 105M DiT and token projection parameters, or 154M including the compressor and decompressor. On LM1B, \JPEGDLM{} entries report $r=0.5$ / $r=0.25$.}
\label{tab:main-uncond}
\centering
\footnotesize
\setlength{\tabcolsep}{3pt}
\renewcommand{\arraystretch}{1.06}
\setlength{\aboverulesep}{1.5pt}
\setlength{\belowrulesep}{1.5pt}
\resizebox{\textwidth}{!}{%
\begin{tabular}{cc*{8}{c}}
\toprule
& & \multicolumn{4}{c}{\textbf{LM1B ($L=128$)}} & \multicolumn{4}{c}{\textbf{OWT ($L=1024$)}} \\
\cmidrule(lr){3-6}\cmidrule(lr){7-10}
\textbf{Method} & \textbf{Param. (M)} & \textbf{Gen-PPL} $\downarrow$ & \textbf{Ent.} & \textbf{MAUVE} $\uparrow$ & \textbf{Speed} $\uparrow$ & \textbf{Gen-PPL} $\downarrow$ & \textbf{Ent.} & \textbf{MAUVE} $\uparrow$ & \textbf{Speed} $\uparrow$ \\
\midrule
AR~\citep{radford2019language} & 130 &  69.06 &  4.394 &  0.952 & \shortstack{$1.00\times$\\($1.62\times$ w/ KV cache)} & 41.06 & 5.583 & 0.930 & \shortstack{$1.00\times$\\($7.14\times$ w/ KV cache)} \\
\midrule
MDLM~\citep{sahoo2024mdlm} & 130 &  115.26 & 4.410 & 0.923 & $0.52\times$ & 41.46 & 5.287 & 0.755 & $0.32\times$ \\
SEDD-Absorb~\citep{lou2024sedd} & 170 & 182.57 & 4.405 & 0.941 & $0.40\times$ & 41.15 & 5.260 & 0.728 & $0.25\times$ \\
Duo~\citep{sahoo2025duo} & 130 &  100.26 & 4.382 & 0.941 & $0.41\times$ & 75.88 & 5.536 & 0.903 & $0.15\times$ \\
FLM~\citep{lee2026flowmap} & 179 &  112.83 & 4.402 & 0.863 & $0.66\times$ & 61.37 & 5.335 & 0.373 & $0.21\times$ \\
LangFlow~\citep{chen2026langflow} & 130 &  92.02 & 4.395 & 0.935 & $0.72\times$ & 36.48 & 5.244 & 0.922 & $0.28\times$ \\
LDLM~\citep{meshchaninov2026ldlm} & 133 (304) & 86.09 & 4.398 & 0.952 & $1.07\times$ & 29.18 & 5.369 & 0.805 & $0.32\times$ \\
ELF-ODE~\citep{hu2026elf} & 105 & 148.76 & 4.344 & 0.948 & $4.01\times$ & 45.50 & 5.307 & 0.842 & $14.50\times$ \\
ELF-SDE~\citep{hu2026elf} & 105 & 117.27 & 4.224 & 0.495 & $4.00\times$ & \textbf{23.93} & 5.156 & 0.711 & $14.49\times$ \\
\midrule
\rowcolor{gray!12}
\textbf{\JPEGDLM{}-ODE (Ours)} & 105 (154) &  74.24 / 88.11 &  4.256 / 4.261 & 0.946 / 0.951 & $6.25\times$ / $\boldsymbol{9.13\times}$ & 34.52 & 5.067 & 0.801 & $\boldsymbol{33.49\times}$ \\
\rowcolor{gray!12}
\textbf{\JPEGDLM{}-SDE (Ours)} & 105 (154) & \textbf{62.07} / \underline{73.03} & 4.232 / 4.237 & 0.915 / 0.913 & $6.24\times$ / \underline{$9.11\times$} & \underline{27.57} & 5.020 & 0.775 & \underline{$33.37\times$} \\
\bottomrule
\end{tabular}
}
\end{table}

\subsection{Analysis of Compressed Latent Representations}
\label{sec:exp-compression}

\begin{figure}[H]
\setlength{\parskip}{0pt}
\centering
\includegraphics[width=\linewidth]{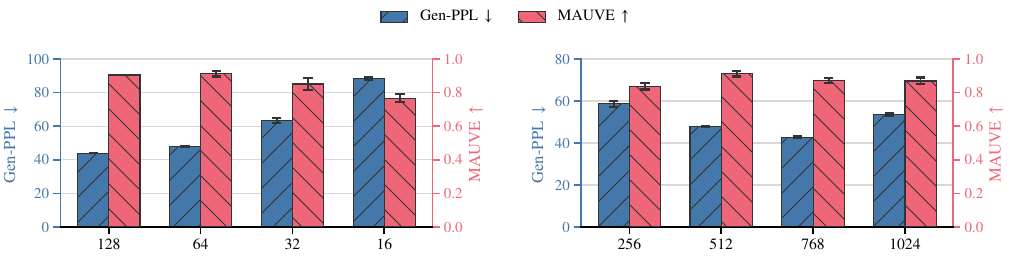}
\par\nointerlineskip
\begin{subfigure}[t]{0.5\linewidth}
\caption{Latent length}
\end{subfigure}%
\begin{subfigure}[t]{0.5\linewidth}
\caption{Latent width}
\end{subfigure}
\caption{Generation quality at different latent lengths (a) and latent widths (b).}
\label{fig:compression-width}
\end{figure}

We study latent-length compression on OWT-128 at $d=512$. Figure~\ref{fig:compression-width}(a) shows that generation quality declines only moderately as compression becomes more aggressive. The modest degradation indicates that \JPEGDLM{} maintains strong generation quality under strong compression.

We next vary the latent width at $r=0.5$. As shown in Figure~\ref{fig:compression-width}(b), increasing the latent width from 256 to 512 substantially improves generation quality. This indicates that a small latent width limits representation capacity. Beyond 512, however, further increasing the latent width does not consistently improve performance.

\subsection{Joint Training vs Two-Stage Training}
\label{sec:exp-cosmos}

In this section, all experiments are conducted on OWT-128. We compare two approaches to generation in a compressed latent space. \JPEGDLM{} jointly trains the compressor, flow matching model, and decoding module. COSMOS first trains an autoencoder and then fits a diffusion model to its fixed embeddings. We retrain COSMOS on the same data with the T5 tokenizer and frozen T5-small encoder used by \JPEGDLM{}. Its compressor and decompressor have the same architectures as ours.

We evaluate \JPEGDLM{} and COSMOS at $r\in\{1.0,0.5,0.25,0.125\}$. Following its default implementation, COSMOS uses final-layer T5-small features at every compression rate. Appendix~\ref{app:cosmos-layer} analyzes its unusually poor result at $r=1.0$. We also train ELF-B~\citep{hu2026elf} on the same data as an uncompressed reference.

\begin{figure}[H]
\vspace{-0.4\baselineskip}
\setlength{\parskip}{0pt}
\centering
\includegraphics[width=\linewidth]{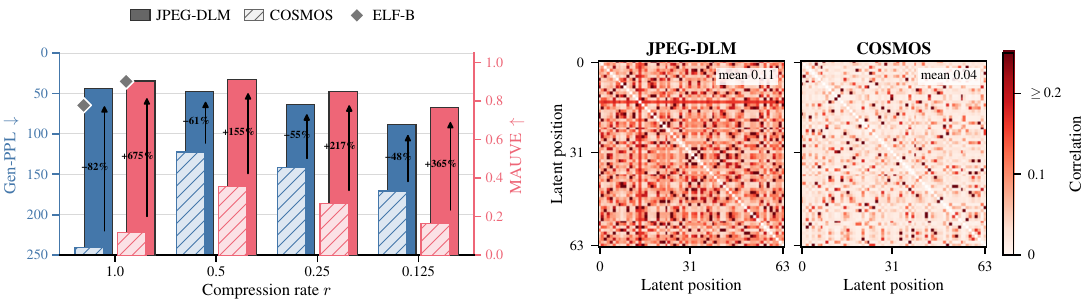}
\par\nointerlineskip
\begin{subfigure}[t]{0.5\linewidth}
\caption{Generation quality}
\end{subfigure}%
\begin{subfigure}[t]{0.5\linewidth}
\caption{Global structure}
\end{subfigure}
\caption{Joint and two-stage training. (a) Gen-PPL and MAUVE across compression rates; arrows show relative changes from COSMOS to \JPEGDLM{}, and diamonds mark ELF-B. COSMOS uses final-layer T5-small features at every rate. (b) Correlation between latent positions at $r=0.5$.}
\label{fig:cosmos-comparison}
\end{figure}

\noindent\textbf{Generation quality.} As shown in Figure~\ref{fig:cosmos-comparison}(a), \JPEGDLM{} achieves lower Gen-PPL and higher MAUVE than COSMOS at every compression rate, using 64 rather than 200 sampling steps. It also achieves lower Gen-PPL than the uncompressed ELF-B reference, which uses 64 steps, at $r\in\{1.0,0.5,0.25\}$. These results show that joint training yields better generation quality at different compression rates.

\vspace{-0.5\baselineskip}

\noindent\textbf{Global structure.} To understand why joint training yields better generation quality, we analyze the global structure of the compressed embeddings. Correlation between latent positions measures how similarly they vary across texts. Higher correlations indicate a clearer structure that is easier for a generative model to learn. At $r=0.5$, we compute pairwise correlations among the 64 positions. Figure~\ref{fig:cosmos-comparison}(b) shows correlations in both embeddings, but the mean off-diagonal correlation is higher for \JPEGDLM{} (0.105) than for COSMOS (0.039). Joint training therefore produces a compressed embedding with stronger global structure, helping explain the better generation quality in Figure~\ref{fig:cosmos-comparison}(a). Further details are provided in Appendix~\ref{app:E3}.

\begin{figure}[t]
\centering
\includegraphics[width=0.56\linewidth]{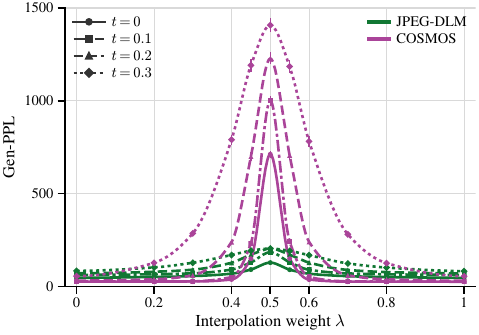}
\caption{Gen-PPL over interpolations between 1{,}024 pairs of clean target embeddings at a compression rate of $r=0.5$, without added noise and at three noise levels. Here, $t$ denotes the COSMOS forward time, with $t=0$ indicating no added noise. For the three noisy settings, \JPEGDLM{} uses noisy embeddings at the same SNRs along its linear path.}
\label{fig:interpolation-ppl}
\end{figure}
\noindent\textbf{Reliable decoding.} During generation, the decoding module receives predicted clean embeddings rather than clean target embeddings. Prediction errors can move these embeddings away from the clean target embedding space, so the decoding module should still decode them into high-quality text. At a compression rate of $r=0.5$, we construct interpolation paths between pairs of clean target embeddings and evaluate each path directly and after adding noise at three strengths. Interpolation traces a controlled path through the embedding space, while the added noise creates progressively larger deviations from clean target embeddings. Figure~\ref{fig:interpolation-ppl} shows that COSMOS decodes high-quality text at the endpoints, but its decoding quality deteriorates sharply near the midpoint. As noise increases, poor decoding extends across a larger part of the interpolation path. In contrast, the decoding quality of \JPEGDLM{} varies smoothly along the path and remains higher near the midpoint at every noise level. These results show that \JPEGDLM{} maintains a broader decodable region between clean target embeddings. This pattern suggests greater tolerance to deviations that can arise in generated compressed embeddings. Further details are provided in Appendix~\ref{app:E3}.

\subsection{Ablation Studies}
\label{sec:exp-ablation}

We ablate the main design choices on OWT-128. All runs use $r=0.5$, latent width $d=512$, a 64-step Euler ODE sampler, and self-conditioning~\citep{chen2023analogbits} at scale $3.0$. The shaded row of Table~\ref{tab:ablation} is the reference point for these studies. Appendix~\ref{app:E3} gives the full OWT-128 configuration.

\noindent\textbf{Shared-weight network.} We share $f_\phi$ between the denoise and decode branches so that flow matching can indirectly shape the compressed embedding space. Retaining JEP but assigning separate networks to the two branches substantially reduces generation quality, showing that weight sharing helps the compressor learn embeddings that are better suited to flow matching.

\noindent\textbf{Two-stage training.} We also train a two-stage variant of \JPEGDLM{} with the same architecture. Stage 1 trains the compressor and the decoding module as an autoencoder on corrupted views with $\mathcal{L}_{\mathrm{feat}}$ and $\mathcal{L}_{\mathrm{ce}}$. It also applies $\mathcal{L}_{\mathrm{jep}}$ with $f_\phi$ in the decode branch as the predictor, so that JEP shapes the compressed embedding space. Stage 2 freezes the compressor and the decoding module and trains only $f_\phi$ with $\mathcal{L}_{\mathrm{flow}}$. The decoding module then decodes the sampled embedding directly. This variant has much higher Gen-PPL and lower MAUVE than Base, which shows that joint training plays a clear role in \JPEGDLM{}.

\noindent\textbf{JEP loss and noise scale.} $\mathcal{L}_{\mathrm{jep}}$ directly trains the context compressor and $f_\phi$ in the decode branch to predict clean target embeddings from corrupted compressed views, shaping a compressed embedding space that supports reliable prediction under corruption. With shared weights, $\mathcal{L}_{\mathrm{jep}}$ and $\mathcal{L}_{\mathrm{flow}}$ use different inputs, but both train the same $f_\phi$ to predict the clean target embedding. Removing JEP alone therefore causes a modest decline because $\mathcal{L}_{\mathrm{flow}}$ continues to train the parameters used by the decode branch to predict clean target embeddings. With separate networks, $\mathcal{L}_{\mathrm{flow}}$ no longer provides this supervision to the decode branch. Removing JEP then causes generation quality to drop sharply, showing that JEP is critical to \JPEGDLM{}.

We further study the corruption noise scale $\sigma$ within JEP. A small scale provides too little pressure to predict clean embeddings robustly, leaving a narrow decodable region. A larger scale exposes the decode branch to larger deviations and can broaden this region, but excessive noise removes information needed to predict the clean embedding. Table~\ref{tab:ablation} shows this balance. The intermediate scale $\sigma=1.0$ performs best, while both $\sigma=0.5$ and $\sigma=1.5$ reduce generation quality.

\noindent\textbf{Feature reconstruction loss.} Token-level cross-entropy supervises text output after token projection, whereas $\mathcal{L}_{\mathrm{feat}}$ directly supervises the decompressed token-length features. Removing $\mathcal{L}_{\mathrm{feat}}$ slows convergence and reduces final generation quality, showing that feature reconstruction stabilizes joint training and improves decoding quality.

\noindent\textbf{Timesteps for the token-level cross-entropy.} We apply $\mathcal{L}_{\mathrm{ce}}$ only to the clean embedding predicted by the decode branch at $t=1$. Extending this loss to embeddings predicted by the denoise branch at $t\in[0.5,1]$ substantially reduces generation quality. These embeddings contain less token-level information, so direct token supervision at intermediate noisy states disrupts the compressed representation. We therefore apply token-level supervision only at $t=1$.

\begin{table}[H]
\caption{Ablation study on OWT-128. Base uses a shared-weight network.}
\label{tab:ablation}
\centering
\footnotesize
\setlength{\tabcolsep}{4pt}
\renewcommand{\arraystretch}{1.00}
\begin{tabular}{lccccccc}
\toprule
\textbf{Variant} & $\mathcal{L}_{\mathrm{jep}}$ & $\mathcal{L}_{\mathrm{feat}}$ & $\sigma$ & $\mathcal{L}_{\mathrm{ce}}$ \textbf{timesteps} & \textbf{Gen-PPL} $\downarrow$ & \textbf{Ent.} & \textbf{MAUVE} $\uparrow$ \\
\midrule
\rowcolor{gray!12}
\textbf{Base}                     & $\checkmark$ & $\checkmark$ & 1.0 & $t=1$ & \textbf{48.00} & 4.064 & \textbf{0.912} \\
two-stage                   & $\checkmark$ & $\checkmark$ & 1.0 & $t=1$ & 91.76 & 4.120 & 0.720 \\
sep. nets                         & $\checkmark$ & $\checkmark$ & 1.0 & $t=1$ & 92.45 & 4.117 & 0.786 \\
w/o $\mathcal{L}_{\mathrm{jep}}$  & $\times$     & $\checkmark$ & 1.0 & $t=1$ & 51.65 & 4.016 & 0.849 \\
w/o $\mathcal{L}_{\mathrm{jep}}$ + sep. nets & $\times$ & $\checkmark$ & 1.0 & $t=1$ & 131.94 & 4.188 & 0.632 \\
w/o $\mathcal{L}_{\mathrm{feat}}$ & $\checkmark$ & $\times$     & 1.0 & $t=1$ & 51.57 & 4.069 & 0.825 \\
$\sigma=0.5$                      & $\checkmark$ & $\checkmark$ & 0.5 & $t=1$ & 61.43 & 3.993 & 0.817 \\
$\sigma=1.5$                      & $\checkmark$ & $\checkmark$ & 1.5 & $t=1$ & 92.78 & 4.046 & 0.545 \\
$\mathcal{L}_{\mathrm{ce}}$ at $t\in[0.5,1]$ & $\checkmark$ & $\checkmark$ & 1.0 & $t\in[0.5,1]$ & 146.80 & 4.140 & 0.645 \\
\bottomrule
\end{tabular}
\end{table}

\section{Conclusion}
\label{sec:conclusion}

We introduced \JPEGDLM{}, which jointly trains a compressor, a flow matching model, and a decoding module to generate text with reduced latent length. JEP shapes the compressed embedding space by training the context compressor and shared-weight network to predict clean target embeddings from corrupted compressed views. Further analyses show stronger global structure and more reliable decoding, supporting our insight that compressed embeddings should be easy for the diffusion model to learn and decode into tokens. \JPEGDLM{} outperforms the two-stage COSMOS baseline at different compression rates and maintains strong generation quality while reaching the highest throughput among diffusion and flow models on LM1B and OWT. These results could motivate further work on jointly learning compressed embeddings for efficient diffusion language modeling.

\clearpage
\section*{AI Use Statement}
Generative AI tools were used to draft and edit parts of the manuscript, suggest the manuscript organization, search for and summarize literature, verify citations and metadata, format references, refine figure presentation, assist with \LaTeX{} formatting, and check the interpretation and presentation of experimental results. All AI-assisted revisions were manually reviewed, cited claims were checked against the original sources, and reported results were verified against the underlying experimental records. We take full responsibility for the final content of this paper, including all text, claims, figures, citations, and artifacts produced with the assistance of generative AI.

\bibliographystyle{iclr2027_conference}
\bibliography{references}

\clearpage
\appendix

\section{Additional Method and Experimental Details}
\label{app:appendix}
\raggedbottom

\setlength{\textfloatsep}{10pt plus 2pt minus 2pt}
\setlength{\floatsep}{10pt plus 2pt minus 2pt}
\setlength{\intextsep}{10pt plus 2pt minus 2pt}
\setlength{\abovecaptionskip}{6pt}
\setlength{\belowcaptionskip}{4pt}
\renewcommand{\arraystretch}{1.06}

\subsection{Method Details}
\label{app:method-details}

\subsubsection{Context and Target Compressors}
\label{app:compressed-target}

Joint-embedding prediction uses a compressed representation of the corrupted view to predict the clean target embedding. We therefore use a context compressor $C_\theta$ to encode the corrupted view and a target compressor $\bar C$ to provide the clean target. The context compressor maps the corrupted view to $\vectorsym{z}_c$ and receives gradients from the JEP objective. The target compressor maps the clean feature matrix $\matrixsym{h}$ to the stop-gradient target $\vectorsym{z}_{\mathrm{tgt}}=\sg\big(\bar C(\matrixsym{h})\big)$ and receives no gradients. The two compressors share the same architecture but follow different update rules.

We update the target compressor as an exponential moving average of the context compressor,
\begin{equation}
\label{eq:app-target-ema}
\bar\theta\leftarrow m\bar\theta+(1-m)\theta,
\end{equation}
where $m$ is the EMA momentum. The stop-gradient fixes the target embedding during each optimization step, while the EMA updates the target compressor from the context compressor. This asymmetric update follows I-JEPA~\citep{assran2023ijepa} and V-JEPA~\citep{bardes2024vjepa}. The momentum schedule for each dataset is reported in Table~\ref{tab:app-system-config}.

\subsubsection{Latent Standardization}
\label{app:latent-regularization}

The compressed embeddings can have different means and scales across channels, while the flow matching path starts from standard Gaussian noise. Similar to COSMOS~\citep{meshchaninov2025cosmos}, we standardize the compressed embeddings before flow matching. This places the clean embeddings and Gaussian noise on comparable scales. We restore the original scale before decoding. Each compressed embedding contains $\Llat$ vectors in $\mathbb{R}^{d}$. During the first part of training, we compute per-channel statistics over all valid latent vectors in each batch. We update running estimates with the momentum reported in Table~\ref{tab:app-system-config} and then fix their final values. Let $\vectorsym{\mu}$ and $\vectorsym{\sigma}$ denote the resulting mean and standard deviation. We apply the same transformation to the outputs of the context and target compressors,
\begin{equation}
\label{eq:app-latent-standardization}
\bar{\vectorsym{z}}=\left(\vectorsym{z}-\vectorsym{\mu}\right)\oslash\vectorsym{\sigma},
\end{equation}
Here $\oslash$ denotes element-wise division. Joint-embedding prediction and flow matching operate on $\bar{\vectorsym{z}}$, and $f_\phi$ predicts standardized embeddings. Before decoding, we invert the transform as $\vectorsym{z}=\bar{\vectorsym{z}}\odot\vectorsym{\sigma}+\vectorsym{\mu}$ and pass the restored embedding to the decompressor during both training and sampling.

\subsubsection{Embedding Regularization}
\label{app:sigreg}

Standardization aligns the mean and scale of each latent channel with Gaussian noise, but it does not control the full latent distribution or prevent channels from losing variation and becoming correlated during joint training. We therefore use Sketched Isotropic Gaussian Regularization (SIGReg) from LeJEPA~\citep{balestriero2025lejepa} and the variance and covariance terms from VICReg~\citep{bardes2022vicreg}. SIGReg encourages an isotropic Gaussian distribution. The variance term maintains variation within each latent channel, and the covariance term reduces redundancy across channels.

Let $U\in\mathbb{R}^{B\times d}$ contain the vectors produced by the context compressor and used to compute the regularizers, with row $\vectorsym{u}_b^\top$. Here $B>1$ is the number of vectors and $d$ is the latent width. SIGReg projects these vectors onto $M$ random unit directions $\vectorsym{a}_m$, sampled uniformly from the unit sphere. For each direction, we compute the empirical characteristic function
\begin{equation}
\label{eq:app-sigreg-ecf}
\widehat\varphi_m(\omega)
=\frac{1}{B}\sum_{b=1}^{B}
\exp\!\left(\mathrm{i}\omega\vectorsym{a}_m^\top\vectorsym{u}_b\right),
\qquad \mathrm{i}^2=-1.
\end{equation}
This function describes the distribution of the projected embeddings. We compare it with the characteristic function $\varphi_0(\omega)=\exp(-\omega^2/2)$ of a standard Gaussian. Using the Epps--Pulley statistic recommended by LeJEPA, the regularizer is
\begin{equation}
\label{eq:app-sigreg-loss}
\mathcal{L}_{\mathrm{sigreg}}
=\frac{B}{M}\sum_{m=1}^{M}
\int_{\mathbb{R}}
\left|\widehat\varphi_m(\omega)-\varphi_0(\omega)\right|^2
w(\omega)\,\mathrm{d}\omega.
\end{equation}
Here $w$ is a Gaussian weight function. LeJEPA's numerical implementation uses $w(\omega)=\exp(-\omega^2/2)$ and trapezoidal quadrature. The loss compares projected distributions beyond their means and variances. Its gradients pass through the projections to the context compressor.

To compute the variance and covariance terms, we first center the vectors and compute their sample covariance,
\begin{equation}
\label{eq:app-embedding-covariance}
\vectorsym{\mu}_U=\frac{1}{B}\sum_{b=1}^{B}\vectorsym{u}_b,
\qquad
\Sigma_U=\frac{1}{B-1}\sum_{b=1}^{B}
(\vectorsym{u}_b-\vectorsym{\mu}_U)
(\vectorsym{u}_b-\vectorsym{\mu}_U)^\top.
\end{equation}
The diagonal entries give the variance of each channel, and the off-diagonal entries measure how channels vary with one another. Following VICReg, we use a hinge penalty on the standard deviations and a squared penalty on the off-diagonal covariances,
\begin{equation}
\label{eq:app-covariance-loss}
\begin{aligned}
\mathcal{L}_{\mathrm{var}}
&=\frac{1}{d}\sum_{i=1}^{d}\max\!\left(0,1-\sqrt{(\Sigma_U)_{ii}+\epsilon}\right),\\
\mathcal{L}_{\mathrm{cov}}
&=\frac{1}{d}\sum_{i\ne j}(\Sigma_U)_{ij}^{2}.
\end{aligned}
\end{equation}
Here $\epsilon>0$ stabilizes the square root. The variance term penalizes channel standard deviations below the unit target used by VICReg. The covariance term penalizes nonzero off-diagonal covariances. Both terms backpropagate to the context compressor.

The full training objective adds these regularizers to the main objective,
\begin{equation}
\label{eq:app-total-objective}
\mathcal{L}_{\mathrm{total}}
=\mathcal{L}_{\mathrm{main}}
+0.1\mathcal{L}_{\mathrm{sigreg}}
+0.2\mathcal{L}_{\mathrm{var}}
+0.002\mathcal{L}_{\mathrm{cov}},
\end{equation}
where $\mathcal{L}_{\mathrm{main}}$ is the four-term objective in Equation~\ref{eq:main-loss}. The target compressor follows the EMA update in Equation~\ref{eq:app-target-ema}. We also ablate the effects of these regularizers in Section~\ref{app:regularizer-ablation}.

\subsubsection{Self-Conditioning}
\label{app:self-conditioning}

During flow matching training, each noisy embedding is sampled independently. During ODE sampling, the model instead predicts a clean embedding at each point along one trajectory. Self-conditioning~\citep{chen2023analogbits} passes the clean embedding predicted at one step to the next, so each step reuses the previous estimate. This technique is widely used in diffusion language models. Following ELF~\citep{hu2026elf}, we also use the previously predicted clean embedding as the conditioning signal for guidance and sample the self-conditioning scale $w$ in the same way. Given a noisy embedding $\vectorsym{z}_t$, we first predict a clean embedding $\widehat{\vectorsym{z}}^{\prime}$ and then condition a second prediction of the clean embedding, $\widehat{\vectorsym{z}}$, on it. We stop gradients through both predicted embeddings and form the training target
\begin{equation}
\label{eq:app-training-self-conditioning}
\begin{aligned}
\widehat{\vectorsym{z}}^{\prime}
&=\sg\big(f_\phi(\vectorsym{z}_t,t;\vectorsym{0},w)\big),\\
\widehat{\vectorsym{z}}
&=\sg\big(f_\phi(\vectorsym{z}_t,t;\widehat{\vectorsym{z}}^{\prime},w)\big),\\
\vectorsym{z}_{\mathrm{SC}}
&=\vectorsym{z}_{\mathrm{tgt}}
+\left(1-\frac{1}{w}\right)
\left(\widehat{\vectorsym{z}}-\widehat{\vectorsym{z}}^{\prime}\right).
\end{aligned}
\end{equation}
The coefficient $1-1/w$ scales the correction $\widehat{\vectorsym{z}}-\widehat{\vectorsym{z}}^{\prime}$ and therefore controls the strength of self-conditioning. At $w=1$, the target reduces to $\vectorsym{z}_{\mathrm{tgt}}$. The final pass uses $\vectorsym{s}=\widehat{\vectorsym{z}}^{\prime}$ and $\vectorsym{z}_{\mathrm{SC}}$ as the target in $\mathcal{L}_{\mathrm{flow}}$. We apply this target with probability $p_{\mathrm{sc}}=0.5$; otherwise, the pass uses $\vectorsym{s}=\vectorsym{0}$ and the clean target $\vectorsym{z}_{\mathrm{tgt}}$. Only the final pass updates the model.

During ODE sampling, the first step uses $\vectorsym{s}_0=0$. Each subsequent step uses the clean embedding predicted at the preceding step,
\begin{equation}
\widehat{\vectorsym{z}}_1^{(k)}
=f_\phi(\vectorsym{z}^{(k)},t_k;\vectorsym{s}_k,w),
\qquad
\vectorsym{s}_{k+1}=\widehat{\vectorsym{z}}_1^{(k)}.
\end{equation}
At inference, each ODE step uses one model evaluation conditioned on $w$. We use $w=3.0$ on LM1B and OWT-128, and $w=2.0$ on OWT-1024.

\subsubsection{\texorpdfstring{$x$-Loss and $v$-Loss for Flow Matching}{x-Loss and v-Loss for Flow Matching}}
\label{app:clean-target-velocity}

The prediction target determines what the flow matching model estimates, while the loss determines how errors at different times are weighted. We consider these choices separately because they affect optimization in different ways. With $x$-prediction, the model directly predicts the clean embedding $\vectorsym{z}_1$. With $v$-prediction, it directly predicts the velocity along the probability path. JiT~\citep{li2026jit} shows that $x$-prediction can be easier for a neural network than $v$-prediction in high-dimensional spaces. We therefore let $f_\phi$ predict $\vectorsym{z}_1$ directly.

For the linear path $\vectorsym{z}_t=t\vectorsym{z}_1+(1-t)\vectorsym{\epsilon}$ in Equation~\ref{eq:flow-path}, differentiating with respect to $t$ gives the target velocity
\begin{equation}
\label{eq:app-conditional-velocity}
\vectorsym{u}_t
=\frac{\mathrm{d}\vectorsym{z}_t}{\mathrm{d}t}
=\vectorsym{z}_1-\vectorsym{\epsilon}.
\end{equation}
Standard flow matching~\citep{lipman2023flowmatching} directly predicts this velocity and minimizes the $v$-loss
\begin{equation}
\label{eq:app-v-loss}
\mathcal{L}_v
=\mathbb{E}_{\vectorsym{z}_1,t,\vectorsym{\epsilon}}
\left[\left\|\vectorsym{v}_\phi(\vectorsym{z}_t,t)-\vectorsym{u}_t\right\|_2^2\right].
\end{equation}
With $x$-prediction, the corresponding $x$-loss is computed directly on the predicted clean embedding,
\begin{equation}
\label{eq:app-x-loss}
\mathcal{L}_x
=\mathbb{E}_{\vectorsym{z}_1,t,\vectorsym{\epsilon}}
\left[\left\|f_\phi(\vectorsym{z}_t,t)-\vectorsym{z}_1\right\|_2^2\right],
\end{equation}
which is the flow matching objective used in Equation~\ref{eq:flow-loss}. An $x$-prediction can be converted into the corresponding velocity prediction for any $t<1$,
\begin{equation}
\label{eq:app-induced-velocity}
\vectorsym{v}_\phi(\vectorsym{z}_t,t)
=\frac{f_\phi(\vectorsym{z}_t,t)-\vectorsym{z}_t}{1-t}.
\end{equation}
Substituting this conversion into the $v$-loss gives
\begin{equation}
\label{eq:app-x-v-loss}
\left\|\vectorsym{v}_\phi-\vectorsym{u}_t\right\|_2^2
=\frac{1}{(1-t)^2}
\left\|f_\phi(\vectorsym{z}_t,t)-\vectorsym{z}_1\right\|_2^2.
\end{equation}
Equation~\ref{eq:app-x-v-loss} weights each $x$-prediction error by $(1-t)^{-2}$ when it is expressed as a velocity error. This weight increases as $t$ approaches $1$. Our logit-normal distribution over $t$ places more probability mass near $t=0$, and the $x$-loss preserves this sampling emphasis. Recent work~\citep{liang2026auroralm} also finds that predicting the clean embedding with an $x$-loss outperforms predicting velocity with a $v$-loss. We therefore use the $x$-loss as the default objective for learning the latent prior with flow matching.

\subsubsection{\texorpdfstring{Relation to Noisy Autoencoding and Consistency Learning}{Relation to Noisy Autoencoding and Consistency Learning}}
\label{app:jep-vs-ae}
Noisy autoencoding supervises only the decoded output. The decoder can compensate for changes in the embedding it receives, so the embedding itself is left unconstrained. JEP adds this constraint. From a corrupted compressed view, $f_\phi$ must recover the clean target embedding, which is the embedding that flow matching models. The w/o $\mathcal{L}_{\mathrm{jep}}$ variant in Table~\ref{tab:ablation} removes this constraint and keeps the corruption, so it serves as a noisy reconstruction baseline. Removing JEP lowers generation quality, and the drop is much larger with separate networks.

A consistency loss trains only the compressor to be invariant to corruption, but generated compressed embeddings never pass through the compressor at inference. JEP also trains $f_\phi$ to recover the clean target embedding from the corrupted compressed view. Generated compressed embeddings pass through $f_\phi$ before decoding, so this robustness is used directly at inference.

\subsection{Additional Results}
\label{app:additional-results}

\subsubsection{Regularizer Ablation}
\label{app:regularizer-ablation}

SIGReg shapes the overall latent distribution, while the variance and covariance terms maintain channel variation and reduce correlations between channels. We therefore evaluate them separately and jointly in Table~\ref{tab:app-regularizer-ablation} to measure their effects on generation quality. The shaded row is the Base configuration from Table~\ref{tab:ablation}.

The runs with only SIGReg and only the variance and covariance terms both have lower Gen-PPL and higher MAUVE than the run without either regularizer. With the variance and covariance terms enabled, adding SIGReg changes entropy from 3.965 to 4.064, MAUVE from 0.868 to 0.912, and Gen-PPL from 40.77 to 48.00. The Base configuration  has the highest MAUVE in this ablation, while the run with only the variance and covariance terms has the lowest Gen-PPL.

\begin{table}[H]
\caption{Regularizer ablation on OWT-128 with $r=0.5$, $d=512$, 64 sampling steps, and self-conditioning at scale $3.0$. The variance and covariance terms are enabled or disabled as a pair. The shaded row reproduces Base from Table~\ref{tab:ablation}.}
\label{tab:app-regularizer-ablation}
\centering
\small
\begin{tabular}{@{}ccccc@{}}
\toprule
\textbf{SIGReg} & \textbf{VICReg} & \textbf{Gen-PPL}$\downarrow$ & \textbf{MAUVE}$\uparrow$ & \textbf{Entropy} \\
\midrule
$\times$ & $\times$ & 63.11 & 0.856 & 4.090 \\
$\checkmark$ & $\times$ & 48.60 & 0.865 & 4.062 \\
$\times$ & $\checkmark$ & \textbf{40.77} & 0.868 & 3.965 \\
\rowcolor{gray!12}
$\checkmark$ & $\checkmark$ & 48.00 & \textbf{0.912} & 4.064 \\
\bottomrule
\end{tabular}
\end{table}

\subsubsection{Generation Quality Across Sampling Steps on LM1B}
\label{app:E4}

To compare the methods across a wider range of sampling budgets, we evaluate 8 to 256 steps on LM1B. The sweep includes \JPEGDLM{} at $r=0.25$ and $r=0.5$, MDLM, Duo, FLM, LangFlow, LDLM, and ELF. We use the checkpoints and scoring from the main comparison. Each point averages five seeds with 1{,}024 samples per seed.

\begin{figure}[H]
\centering
\includegraphics[width=0.76\textwidth]{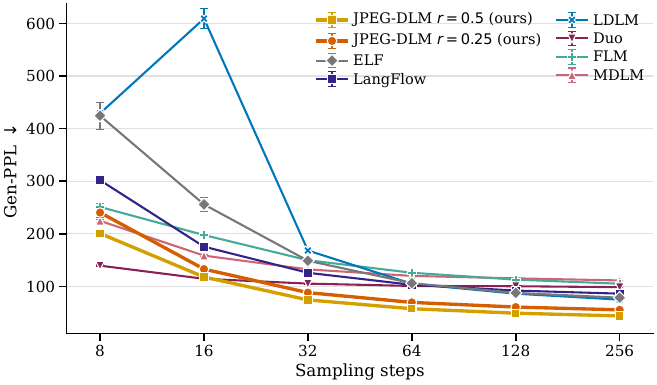}
\caption{Gen-PPL on LM1B across sampling steps. Error bars show standard deviations over five seeds.  \JPEGDLM{} is shown at $r=0.5$ and $r=0.25$.}
\label{fig:lm1b-step-sweep}
\end{figure}

Figure~\ref{fig:lm1b-step-sweep} shows that \JPEGDLM{}  has lower Gen-PPL than every baseline from 32 steps onward, at both compression rates. Its Gen-PPL decreases steadily with more steps.

\begin{table}[H]
\caption{Generation quality of \JPEGDLM{} on LM1B  at $r=0.5$ and $r=0.25$ at different ODE sampling budgets. Results are means over five seeds. The shaded row is the 32-step configuration used in the main comparison.}
\label{tab:app-lm1b-steps}
\centering
\footnotesize
\setlength{\tabcolsep}{4pt}
\renewcommand{\arraystretch}{1.06}
\begin{tabular}{c*{6}{c}}
\toprule
& \multicolumn{3}{c}{\textbf{2$\times$ compression} ($r=0.5$)} & \multicolumn{3}{c}{\textbf{4$\times$ compression} ($r=0.25$)} \\
\cmidrule(lr){2-4}\cmidrule(lr){5-7}
\textbf{Steps} & \textbf{Gen-PPL} $\downarrow$ & \textbf{Ent.} & \textbf{MAUVE} $\uparrow$ & \textbf{Gen-PPL} $\downarrow$ & \textbf{Ent.} & \textbf{MAUVE} $\uparrow$ \\
\midrule
8   & $200.64$ & $3.952$ & $0.094$ &  $240.38$ &  $4.114$ &  $0.338$ \\
16  & $117.85$ & $4.249$ & $0.882$ &  $132.85$ &  $4.269$ &  $0.912$ \\
\rowcolor{gray!12}
32  & $74.24$ & $4.256$ & $\mathbf{0.946}$ &  $88.11$ &  $4.261$ & $\mathbf{0.951}$ \\
64  & $57.52$ & $4.238$ & $0.935$ &  $69.53$ &  $4.244$ &  $0.944$ \\
128 & $48.98$ & $4.223$ & $0.928$ &  $60.65$ &  $4.229$ &  $0.938$ \\
256 & $\mathbf{43.83}$ & $4.209$ & $0.911$ &  $\mathbf{55.40}$ &  $4.220$ &  $0.940$ \\
\bottomrule
\end{tabular}
\end{table}

\subsubsection{\texorpdfstring{Gen-PPL and Entropy Across Self-Conditioning Scales on OWT-1024}{Gen-PPL and Entropy Across Self-Conditioning Scales on OWT-1024}}
\label{app:sc-sweep}
\begin{figure}[H]
\centering
\includegraphics[width=0.5\textwidth]{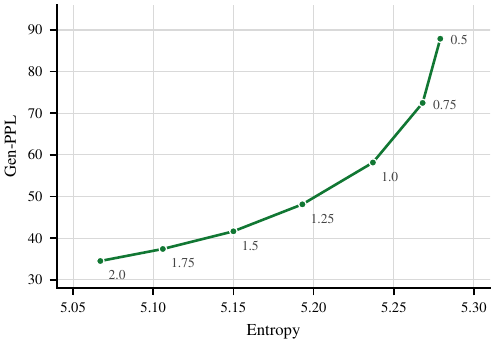}
\caption{Gen-PPL--entropy trade-off of \JPEGDLM{} on OWT-1024. Increasing the self-conditioning scale improves Gen-PPL but reduces entropy. Labels mark the scale of each point.}
\label{fig:app-sc-sweep}
\end{figure}

Following ELF~\citep{hu2026elf}, we evaluate the Gen-PPL--entropy trade-off of \JPEGDLM{} on OWT by varying the self-conditioning scale, which controls the balance between generation quality and diversity (Figure~\ref{fig:app-sc-sweep}).

\subsection{Limitations and Future Work}
\label{app:limitations}

\noindent\textbf{Limitations.} First, although \JPEGDLM{} matches or outperforms the uncompressed diffusion and flow baselines in Gen-PPL in short-text generation while using latent-length compression, this advantage does not yet extend uniformly to long-text generation. On OWT, with twofold latent-length compression, \JPEGDLM{} does not yet outperform every uncompressed baseline across all quality measures (Table~\ref{tab:main-uncond}). Nevertheless, Figure~\ref{fig:cosmos-comparison} shows that \JPEGDLM{} consistently outperforms the existing two-stage compressed baseline across compression rates. These results suggest that jointly learning compressed embeddings is a promising direction for text generation under latent-length compression.

Second, the corruption noise scale is a training hyperparameter that requires careful calibration. It controls an interpretable balance between robust prediction and information preservation. The ablation in Table~\ref{tab:ablation} follows this expected pattern. An intermediate scale produces the best overall quality. Weaker corruption provides insufficient pressure to predict clean embeddings robustly, whereas excessive corruption removes information needed to recover the clean target embedding.

\noindent\textbf{Future work.} Our results show that learning compressed embeddings jointly with the generative model yields better generation quality than two-stage training at different compression rates. Unlike two-stage training, joint training allows the embedding space to adapt indirectly to the generative objective while it is still trained for token decoding. The goal is therefore to learn compressed embeddings with a structure that is easy for the diffusion model to learn and allows generated embeddings to be decoded reliably into tokens. Accordingly, we consider three directions for improving text generation quality under stronger latent-length compression, and one direction for further reducing the sampling cost.

\noindent\textbf{Compressor architecture.} The current compressor and decompressor attend over their complete inputs, so compressed positions have no fixed correspondence with token spans. AURORA-LM~\citep{liang2026auroralm} instead uses a query-based encoder with causal attention to form a prefix-aligned latent sequence. Its results suggest that causal encoding is a promising direction for retaining positional structure as latent length decreases.

\noindent\textbf{Contextual token decoding.} The current token projection predicts each token independently from its decompressed feature and does not model dependencies among output tokens. CoDAR~\citep{shen2026codar} finds that an autoregressive Transformer decoder maps continuous representations back to tokens more accurately than a position-wise linear head. A contextual decoder may therefore improve decoding under stronger compression.

\noindent\textbf{Alternative joint training objectives.} JEP is one way to shape the compressed embedding space during joint training. Another approach is to update the compressor directly with the generative loss at each sampled time step. LDLM~\citep{meshchaninov2026ldlm} follows this strategy and uses several techniques to stabilize training. In our preliminary experiments, applying flow matching gradients to the compressor at each sampled time step was difficult to stabilize and led to representation collapse. Further work on training objectives and stabilization may yield a compressed embedding space that is easier for the diffusion model to learn. This could improve text generation quality under stronger compression.

\noindent\textbf{Few-step distillation.} Distillation methods such as FMLM~\citep{lee2026flowmap} and DFM~\citep{potaptchik2026dfm} reduce the number of sampling steps, whereas \JPEGDLM{} reduces the latent length. The two approaches are complementary, and applying few-step distillation to \JPEGDLM{} is a direction for future work.

\subsection{Experimental Details}
\label{app:experimental-details}

Table~\ref{tab:app-system-config} lists the model configurations and training schedules for LM1B and OWT-1024. The OWT-128 comparisons and ablations are detailed in Appendix~\ref{app:E3}.

\subsubsection{Data Preparation}
\label{app:E1}

We follow the packed LM1B preprocessing of Duo~\citep{sahoo2025duo}, adapted for the frozen T5 encoder. For OWT-1024 we pack text  into blocks of exactly 1{,}024 T5 tokens that start at sentence boundaries, so every position holds a real token and no padding is needed. Packing makes EOS a document boundary rather than a sequence terminator, so we keep the text after the first EOS when scoring.

\subsubsection{Truncated MAUVE}
\label{app:mauve-diagnostic}

\begin{figure}[!t]
\centering
\includegraphics[width=\textwidth]{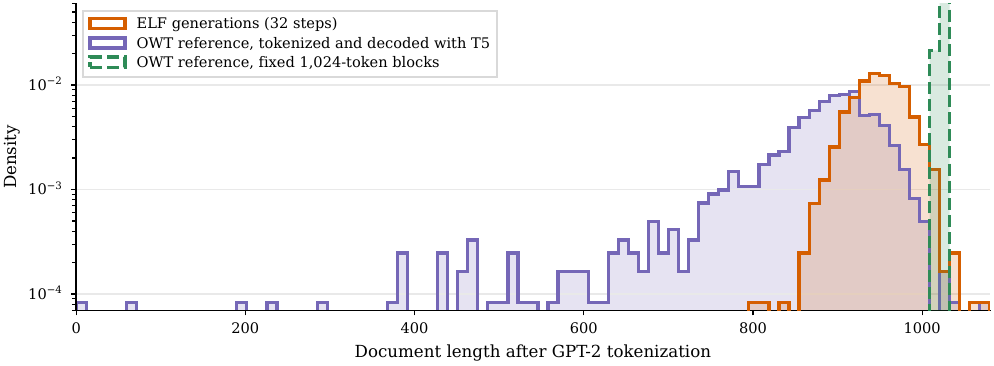}
\caption{GPT-2 token lengths of ELF generations and two OWT reference sets. The reference constructed from fixed 1{,}024-token blocks concentrates near the block length, whereas the reference tokenized and decoded with T5 and the ELF samples vary in length.  The ELF and fixed-length reference histograms  overlap by only 0.9\%.}
\label{fig:app-mauve-length}
\end{figure}

For MAUVE~\citep{pillutla2021mauve}, we represent each text with the hidden state at its final retained GPT-2 token. Because this state is extracted at the end of the retained context, the resulting representation varies with text length and boundary placement as well as language quality. Consequently, differences in length distribution can confound the comparison between generated and reference texts. For OWT-1024, we construct one reference set by splitting a GPT-2 token stream into fixed blocks of 1{,}024 tokens. ELF produces its samples differently. It generates on a 1{,}024-position T5 canvas and decodes with T5 SentencePiece. After GPT-2 re-tokenization, its outputs therefore follow a different length distribution. Figure~\ref{fig:app-mauve-length} also shows a second OWT reference set that was first tokenized with T5 and then decoded.

This construction difference leads to a large length mismatch, as shown in Figure~\ref{fig:app-mauve-length}. For 32-step ELF samples, the generated documents have a median length of 950 GPT-2 tokens and a standard deviation of 31.9. By comparison, the fixed-length reference has a median length of 1{,}021 tokens and a standard deviation of 1.0. The two length histograms overlap by only 0.9\%. Consequently, MAUVE computed on the full texts can reflect the length mismatch in addition to differences in generated text.

TEncDM~\citep{shabalin2025tencdm} reports the same sensitivity on ROCStories. It truncated both generated and reference texts to 40 tokens. After truncation, MAUVE increased from 0.762 to 0.940 with BERT encodings and from 0.647 to 0.913 with RoBERTa encodings. Thus, text length can substantially change the MAUVE score. Accordingly, we report truncated MAUVE by truncating generated and reference texts to the same GPT-2 token budget before feature extraction. As a result, the final retained token appears at the same position for nearly all texts, which reduces the influence of length on the resulting representation. We use a budget of 96 tokens on LM1B and OWT-128 and 768 tokens on OWT-1024. Both budgets retain 75\% of the nominal sequence length.

\subsubsection{Model and Training Configuration}
\label{app:jpeg-config}

The latent DiT follows ELF-B~\citep{hu2026elf}, with separate output heads for the denoise and decode branches. It prepends four time tokens, four self-conditioning tokens, and four model-mode tokens to the latent sequence.

\begin{table}[!t]
\caption{Model and training configurations of \JPEGDLM{} on LM1B and OWT-1024. Sampling compute counts one multiply--accumulate as one operation. For $K$ solver steps, \JPEGDLM{} uses $K+1$ DiT evaluations, including the final clean embedding prediction, followed by one decoding pass. The estimate at full latent length uses the same $K+1$ evaluation budget.}
\label{tab:app-system-config}
\centering
\footnotesize
\renewcommand{\arraystretch}{0.98}
\begin{tabular}{@{}>{\raggedright\arraybackslash}p{0.37\textwidth}>{\raggedright\arraybackslash}p{0.27\textwidth}>{\raggedright\arraybackslash}p{0.27\textwidth}@{}}
\toprule
\textbf{Parameter} & \textbf{LM1B} & \textbf{OWT-1024} \\
\midrule
\multicolumn{3}{@{}l}{\emph{Data and tokenization}} \\
Dataset & LM1B~\citep{chelba2014one} & OWT \\
Token sequence length & 128 & 1024 \\
Sequence construction & Blocks packed across sentence boundaries & Blocks packed across sentence boundaries (no padding) \\
\midrule
\multicolumn{3}{@{}l}{\emph{Model architecture}} \\
Frozen text encoder & T5-small & T5-small \\
Latent length / compression rate $r$ & 64 / 0.5, 32 / 0.25 & 512 / 0.5 \\
Latent width & 512 & 512 \\
Compressor / decompressor layers & 6 / 6 & 6 / 6 \\
Compressor attention heads & 8 & 8 \\
Feature bottleneck & 128 & 128 \\
DiT layers / hidden size / heads & 12 / 768 / 12 & 12 / 768 / 12 \\
DiT MLP ratio / dropout & 4 / 0 & 4 / 0 \\
DiT + token projection parameters & 105M & 105M \\
Including compressor and decompressor & 154M & 154M \\
\midrule
\multicolumn{3}{@{}l}{\emph{Training}} \\
Training epochs &  30 & 5 \\
Number of GPUs & 8 & 8 \\
Local batch / gradient accumulation & 384 / 2 & 24 / 2 \\
Effective global batch size & 6144 & 384 \\
Optimizer & AdamW ($\beta_1{=}0.9$, $\beta_2{=}0.999$) & AdamW ($\beta_1{=}0.9$, $\beta_2{=}0.999$) \\
Precision & bfloat16 & bfloat16 \\
Gradient clipping / weight decay & 1.0 / 0.01 & 1.0 / 0.01 \\
Flow matching learning rate & $1\times10^{-3}$ & $1\times10^{-3}$ \\
Compressor / decompressor learning rate & $3.0\times10^{-4}$ & $2\times10^{-4}$ \\
Feature perturbation & Gaussian, $\sigma=0.8$ & Gaussian, $\sigma=0.8$ \\
Flow matching / JEP / token / feature weights & 1 / 1 / 1 / 1 & 1 / 1 / 1 / 1 \\
SIGReg weight & 0.1 & 0.1 \\
Latent standardization momentum & 0.99 & 0.99 \\
EMA decay for target compressor & $0.995\rightarrow0.999$ &  $0.995\rightarrow0.999$ \\
Model weight EMA & 0.999 & 0.999 \\
Time sampling & Logit-normal $(-1.5,0.8)$ & Logit-normal $(-1.5,0.8)$ \\
Self-conditioning probability & 0.5 & 0.5 \\
Sampling compute (compressed / full latent length) & 0.13 ($r=0.25$) / 0.40~TFLOPs & 1.68 / 3.56 TFLOPs \\
\bottomrule
\end{tabular}
\end{table}

\subsubsection{Baseline Details}
\label{app:baseline-config}

Each model uses the tokenizer it was trained with, as listed in Table~\ref{tab:app-baseline-config}. The AR baseline uses nucleus sampling~\citep{holtzman2020nucleus}.

On OWT, we use the official checkpoints of MDLM~\citep{sahoo2024mdlm}, SEDD-Absorb~\citep{lou2024sedd}, Duo~\citep{sahoo2025duo}, FLM~\citep{lee2026flowmap}, LangFlow~\citep{chen2026langflow}, and ELF~\citep{hu2026elf}, and the reference AR checkpoint released with MDLM. On LM1B, we use the official FLM and LangFlow checkpoints, and the MDLM and Duo checkpoints released with FLM. For LDLM~\citep{meshchaninov2026ldlm}, we use its official checkpoints on both datasets. AR, SEDD-Absorb, and ELF have no officially released LM1B weights, so we train these models ourselves following their official recipes.

\begin{table}[!t]
\caption{Tokenization and sampling configurations for the main unconditional generation comparison. BERT~\citep{devlin2019bert} denotes \texttt{bert-base-uncased}, and GPT-2~\citep{radford2019language} denotes GPT-2 BPE. T5~\citep{raffel2020t5} uses SentencePiece~\citep{kudo2018sentencepiece}. A checkmark in the SC column indicates self-conditioning.}
\label{tab:app-baseline-config}
\centering
\small
\renewcommand{\arraystretch}{1.08}
\setlength{\tabcolsep}{9pt}
\begin{tabular}{@{}l c c l c l c@{}}
\toprule
& & & \multicolumn{2}{c}{\textbf{LM1B}} & \multicolumn{2}{c}{\textbf{OWT-1024}} \\
\cmidrule(lr){4-5}\cmidrule(l){6-7}
\textbf{Method} & \textbf{SC} & \textbf{Samples} & \textbf{Tokenizer} & \textbf{Steps} & \textbf{Tokenizer} & \textbf{Steps} \\
\midrule
AR & $\times$ & 1{,}024 & BERT & -- & GPT-2 & -- \\
\midrule
MDLM & $\times$ & 1{,}024 & BERT & 128 & GPT-2 + mask & 1024 \\
SEDD-Absorb & $\times$ & 1{,}024 & GPT-2 & 128 & GPT-2 & 1024 \\
Duo & $\times$ & 1{,}024 & BERT & 128 & GPT-2 & 1024 \\
\midrule
FLM & $\times$ & 1{,}024 & BERT & 128 & GPT-2 & 1024 \\
LangFlow & $\checkmark$ & 1{,}024 & BERT & 128 & GPT-2 & 1024 \\
LDLM & $\checkmark$ & 1{,}024 & GPT-2 & 128 & GPT-2 & 1024 \\
ELF & $\checkmark$ & 1{,}024 & T5 & 32 & T5 & 32 \\
\midrule
\rowcolor{gray!12}
\textbf{\JPEGDLM{} (ours)} & $\checkmark$ & 1{,}024 & T5 & 32 & T5 & 32 \\
\bottomrule
\end{tabular}
\end{table}

\subsubsection{Evaluation Details}
\label{app:evaluation-details}

For the main comparison, each model uses its own sampler and the step count in Table~\ref{tab:app-baseline-config}. For \JPEGDLM{}, we draw the initial embedding from a standard Gaussian and integrate the learned velocity field with a 32-step Euler solver. The decode branch then applies $f_\phi(\cdot,1)$ once, restores the predicted clean embedding to its original scale, and maps it to T5 tokens with greedy decoding. Because the models use different vocabularies, we decode each sample to text and re-tokenize it with GPT-2 Large before scoring. We also lowercase the LM1B outputs of ELF to match the uncased reference. For the LM1B baselines that use the BERT tokenizer, we follow their official evaluation and keep the \texttt{[CLS]} and \texttt{[SEP]} tokens when computing Gen-PPL. These tokens do not appear in the reference text and can lower MAUVE, so we remove them when computing MAUVE. Gen-PPL is the exponential of the average negative log-likelihood over all scored tokens, and unigram entropy is reported in nats.

We measure median generation throughput in FP32 at batch 64 on LM1B and batch 16 on OWT-1024. The timing excludes model loading and detokenization. AR without a KV cache defines the $1.00\times$ baseline. With a KV cache, AR reaches $1.62\times$ this throughput on LM1B and $7.14\times$ on OWT-1024.

\subsubsection{Experiments on OWT-128}
\label{app:E3}

\noindent\textbf{Training configuration.}
We use the \nolinkurl{linluqiu/openwebtext-len128-packing_padding} dataset, whose sequences contain 128 T5 tokens. These experiments use the OWT-1024 architecture and optimization settings in Table~\ref{tab:app-system-config}. For OWT-128, we train for 3 epochs and use $\sigma=1.0$, 64 sampling steps, and self-conditioning scale $3.0$. The studies below specify the compression rate and latent width that they vary.

\noindent\textbf{Latent compression and component ablations.}
The latent-length study fixes $d=512$ and evaluates $r\in\{1.0,0.5,0.25,0.125\}$. The latent-width study fixes $r=0.5$ and evaluates $d\in\{256,512,768,1024\}$. Figure~\ref{fig:compression-width} reports these results. The component ablations in Table~\ref{tab:ablation} use $r=0.5$ and $d=512$. They evaluate weight sharing, joint-embedding prediction, feature reconstruction, noise scale, and cross-entropy timesteps. The two-stage variant follows the same setting as the other ablations. Stage 1 trains the compressor, the decoding module, and $f_\phi$ as the JEP predictor for 1 epoch, and the decoding module learns from the compressed embedding of the corrupted view rather than from the prediction of $f_\phi$. Stage 2 initializes $f_\phi$ from Stage 1, freezes the compressor and the decoding module, and trains $f_\phi$ with $\mathcal{L}_{\mathrm{flow}}$ for 2 epochs. At inference, the frozen decoding module decodes the sampled embedding directly.

\noindent\textbf{Baselines.}
We retrain the T5 variant of COSMOS with the same frozen T5-small encoder, tokenizer, and compressor and decompressor depths as \JPEGDLM{}. Table~\ref{tab:app-cosmos-config} lists its configurations across compression rates. ELF-B uses its official hyperparameters on this dataset and serves as the uncompressed baseline.

\begin{table}[!t]
\caption{Configuration of COSMOS on OWT-128.}
\label{tab:app-cosmos-config}
\centering
\small
\renewcommand{\arraystretch}{1.08}
\begin{tabular}{@{}p{0.42\textwidth}p{0.53\textwidth}@{}}
\toprule
\textbf{Parameter} & \textbf{OWT-128} \\
\midrule
Dataset & \nolinkurl{linluqiu/openwebtext-len128-packing_padding} \\
Sequence length & 128 \\
Text encoder / tokenizer & Frozen T5-small / T5 \\
Encoder feature & Final hidden layer \\
Latent length / compression rate $r$ & 128 / 1.0, 64 / 0.5, 32 / 0.25, 16 / 0.125 \\
Latent width & 512 \\
Compressor / decompressor layers & 6 / 6 \\
Attention heads & 8 \\
Vocabulary size & 32{,}100 \\
\leavevmode Sampling steps / schedule parameter & \leavevmode200 / 5 \\
Token masking / MLM probability & 0.5 / 0.3 \\
Gaussian noise weight / scale & 0.5 / 0.7 \\
Latent masking probability & 0.4 \\
\bottomrule
\end{tabular}
\end{table}

\noindent\textbf{COSMOS performance at $r=1.0$.}
\label{app:cosmos-layer}
Following its default implementation, COSMOS feeds the final hidden layer of the frozen T5-small encoder to the compressor. We denote this feature as layer $-1$. With this feature, COSMOS performs worse at $r=1.0$ than at $r=0.5$, even though $r=1.0$ retains one latent position per token. We examine whether the encoder feature contributes to this result by training additional $r=1.0$ models using hidden states from layers $-2$ and $-3$.

\begin{table}[H]
\caption{COSMOS performance at $r=1.0$ with T5-small layer indices $-1$, $-2$, and $-3$. Results are means over five seeds using 1{,}024 samples per seed.}
\label{tab:app-cosmos-layer}
\centering
\footnotesize
\setlength{\tabcolsep}{5pt}
\renewcommand{\arraystretch}{1.08}
\begin{tabular}{@{}lccc@{}}
\toprule
\textbf{Layer index} & \textbf{Gen-PPL} $\downarrow$ & \textbf{Ent.} & \textbf{MAUVE} $\uparrow$ \\
\midrule
$-1$ & $240.81$ & $3.903$ & $0.117$ \\
$-2$ & $154.55$ & $3.974$ & $0.301$ \\
$-3$ & $163.20$ & $3.972$ & $0.268$ \\
\bottomrule
\end{tabular}
\end{table}

The poor result at $r=1.0$ cannot be explained by limited latent capacity, because this configuration retains one latent position per token. COSMOS trains its autoencoder before the diffusion model, so the diffusion model must fit the latent space induced by the selected T5 feature. \mbox{LDLM~\citep{meshchaninov2026ldlm}} reports that the encoder layer has a large effect on a COSMOS-style latent space. The final layer gives substantially worse MAUVE than earlier layers. LDLM suggests that the smoothing and robustness objectives are sensitive to the structure of the pretrained representation.

Our results show the same pattern with T5-small. Changing the input from layer $-1$ to layer $-2$ lowers Gen-PPL from 240.81 to 154.55 and raises MAUVE from 0.117 to 0.301. Layer $-3$ reaches a Gen-PPL of 163.20 and a MAUVE score of 0.268. It is slightly worse than layer $-2$ but remains much better than layer $-1$. This comparison shows that the final T5 feature accounts for much of the unusually poor result at $r=1.0$. However, changing the encoder layer does not close the gap. The $r=0.5$ model with layer $-1$ reaches a Gen-PPL of 122.46 and a MAUVE score of 0.358. At $r=0.5$, each latent position must aggregate information from multiple token features. This stronger bottleneck reduces token-level variation and imposes more structure on the latent representation. The remaining gap is therefore consistent with compression making the latent distribution easier for the diffusion model to learn. Figure~\ref{fig:cosmos-comparison}(a) follows the default COSMOS implementation and reports layer $-1$ at every compression rate. The results from layers $-2$ and $-3$ are used only for this diagnostic. At $r=0.5$, the final-layer feature performs best among the T5 layers we tested for COSMOS.

\noindent\textbf{Generation quality.}
For Figure~\ref{fig:cosmos-comparison}(a), COSMOS uses 200 sampling steps, while \JPEGDLM{} and ELF-B use 64. We re-tokenize the decoded samples with GPT-2 Large to compute Gen-PPL and unigram entropy. MAUVE uses 1{,}024 OWT reference documents and the first 96 GPT-2 tokens from each generated and reference text.

\noindent\textbf{Global structure.}
We sample 2{,}048 sequences from distinct OWT documents and remove duplicate token sequences. We encode them with frozen T5-small and apply each compressor at $r=0.5$. The \JPEGDLM{} target compressor and the COSMOS autoencoder compressor produce clean compressed embeddings with 64 latent positions and width 512.

We transform the embeddings into the standardized coordinates used by each generative model. At each latent position, we subtract the mean across sequences. We then flatten each position across sequences and channels, normalize the resulting 64 vectors, and compute their $64\times64$ Gram matrix. Figure~\ref{fig:cosmos-comparison}(b) displays the off-diagonal entries with a shared color scale.

\noindent\textbf{Reliable decoding.}
We form 1{,}024 random pairs of clean compressed embeddings. For each pair $(\vectorsym{z}_a,\vectorsym{z}_b)$, we compute
\begin{equation}
\vectorsym{z}_{\lambda}=(1-\lambda)\vectorsym{z}_a+\lambda\vectorsym{z}_b,
\qquad
\lambda\in\{0,0.2,0.3,0.4,0.45,0.5,0.55,0.6,0.7,0.8,1\}.
\end{equation}
We interpolate in the standardized coordinates used by each generative model. Without added noise, we invert each model's standardization and pass $\vectorsym{z}_{\lambda}$ to its decoding module without first applying the generative model. The three noise levels use COSMOS forward times $t\in\{0.1,0.2,0.3\}$ with $\operatorname{SNR}(t)=[5\tan(\pi t/2)]^{-2}$. For \JPEGDLM{}, the times $t'\in\{0.5581,0.3810,0.2819\}$ on the linear path match these three SNR values.

We draw Gaussian noise and normalize each tensor to unit RMS. A shared noise tensor is applied to both methods at every $\lambda$. At each noisy state, the generative model predicts one clean embedding, which is restored to the original latent scale and mapped to text. For this diagnostic, \JPEGDLM{} uses self-conditioning scale 1.0, and COSMOS receives a zero self-conditioning input. We compute Gen-PPL with GPT-2 Large for the 1{,}024 decoded texts at each $(\lambda,t)$ pair. Figure~\ref{fig:interpolation-ppl} reports 95\% intervals from 2{,}000 bootstrap resamples over texts and connects the measured values with monotone cubic interpolation.

\clearpage
\subsection{Qualitative Generation Samples}
\label{app:qualitative}

We include two unconditional generations produced by the \JPEGDLM{} configuration used for the OWT-1024 results in Table~\ref{tab:main-uncond}.

\nolinenumbers
\begin{generationexample}{1}
\fontsize{8pt}{9pt}\selectfont
But the fact that ESPN has proven to be the most successful news company in the United States is far more complicated than the nature of it. The reason why ESPN is changing, as Mr. Sargent asserted, is about what happens in the US. While Mr. Warner has not commented with ESPN’s mediocry pay-and-ftar business, his comments have received widespread criticism from some executives, many who don’t have a grasp of what is most important for him. It’s a reflection of ESPN’s business culture.

For nearly 20 years, Mr. Fox has been on the radar of prominent executives and broadcast executives, conducting face-to-face interviews with the San Angeles-based NPR network. In 2015, he made his relationship with Richie Fitina, the company’s mosttime content officer, to ESPN’s \$1 billion-million video streaming service, Newstime. Sargent said he is not aware of Mr. Warner’s comments as a sign of personal bias, but that he has spent many time working on media shows. “I’ve never seen any time in my career that no person or anyone in a media company has been in the US,” he said. While Sargent hasn’t seen ESPN’s own news media legecies yet, he says that’s not a good thing. Its motivations are driven on the nature of ESPN’s relatively established media brand. “My pitch-and-feek partner, John J. Sargent has brought me to the point of what he really cares to me at ESPN,” Sargent adds. “I think that’s not in the same way that ESPN has,” he said. “I think that’s my own opinion, and I really believe with that.”

As a journalist, Sargent argues that ESPN’s arrogive, sknical view of the nature of its media business has made him more likely to carry out its own TV shows with no much scrutiny. When he gave a deal with ESPN in September in late 2014, Sargent’s comments sparked widespread criticism and outsternation about the company’s leadership. Those claims are false. Authors like Foxcasts.com’s Cook, who is ESPN’s VP producer and chief director for FoxWorks, and Fox Warner Inc.’s David Smith of Hollywood have dismissed the criticism of Mr. Warner’s comments, Sargent said. But those claims aren’t true. “I understand the validity of these claims,” Sargent said. “But I think it very important is that if you watch shows like ESPN and Showtime on a very large basis, people will get away with them. And if you’re a journalist, I think that is correct.”

A key concern is that ESPN is based on its brand, rather than content. For some people, ESPN says it has very more control with its content model — that it wants its subscribers to watch shows they show on on their own channels. However, its criticisms, unsubstantsive, have nothing to do with broadcast viewers’ behavior, as the company’s ability to include content on their channels would encourage viewers to sign up to their shows. And despite Mr. Warner’ claims, there is also skepticism that ESPN’s content strategy — like Mr. Warner’s work with Fasttime — will be difficult to ignore. Sargent’s argument is that ESPN will offer more content to its subscribers on its channel, as opposed to keeping them out at the top of their TV list. But the rise of social media and streaming platforms has made it more likely that ESPN will deliver more content on its channels if it can. For many people, this is what ESPN is doing now.

It’s a two-year problem, and has led to a sharp increase in criticism from executives of all stripes. In an interview post, John Benzee ounced ESPN’s relationship to ESPN’s upcoming programs. Last month, Benzee lagrily accused Mr. Warner of making comments — denouing the company’s plan to launch a new service, Newstime — “unappropriable.” Benzee refused to make questions about ESPN’s new service, Newstime and its new Newstime service, aiming to save the company’s so-called content to be more accessible to other audiences. “Let’s not forget that no one in the United States is happy about this now,” said David Barey, University of Chicago’s Department of Business, Media, and Politics. “It concerns me of the importance of ESPN and his people of all the world and the fact that he’s still at the top of the line.”
\end{generationexample}

\clearpage
\begin{generationexample}{2}
\fontsize{8pt}{9pt}\selectfont
Even though Windberg’s statements are correct, this week’s decision will show that that’s consistent with the laws we’re going to make in California,” California Publicity Chairman John Russler said. “Bushberg’s statements were not accurate,” in a statement. “Bushberg had said he had violate to Section 230 of the Bill of Rights, but the court was satisfied that [Bushberg] had held violation of his law.” The ruling, which will be heard in a California court court, is to begin in month.

On Friday the U.S. Court of Appeals spelled a court order pending an order to halt Windway, an organization whose content is non-ivo content. Gov. Mike Schugan is slated to speak before the Supreme House on it on Dec. 1. In a statement on Tuesday, Schuich said the state would block sale of the news company’s business as a marketing service to California-based Fastfeed Media, an a media company founded in 2005 and founded by Rida Bardo. California Gov. Scott Scott said the court’s decision didn’t change Windberg’s claims. “When I’m elected, I’m disgusted,” he said. The state has refused to block the sale of Winddoor’s advertising service to Amazon.

“This decision clearly has a serious impact to the lives of people who want to share our media content. I am deeply appalened by it, and I believe that it would be contrary of the constitutional principles of the federal Amendment,” Windway CEO John Gushberg said in a written statement. Although Windwater shifted its business to Amazon, it also shifted its business by promoting a paid service. In June, Windwater decided to cease its business.

In a decision issued Monday by the U.S. Department of Justice, the California Court of Appeals said Winddoor had suspended its business. The news company viewed more than 100 videos posted on YouTube, including videos like “FikeBids” on a wide range of topics, including movies, “Bake,” and “Bakeball,” David Smith, one of Windway’s founders at Goldman Networks Inc., argued in a statement. Earlier Tuesday, Michael J. Saffer, who turned out Windberg’s opinion in the ruling, said that prosecutors would now intend to issue an order on Winddoor’s business, which violated of the Freedom of Information Act and the Freedom of Information Act. “Today, the government is going to ensure that all consumers and consumers will be protected by this decision.”

Jared Suly, a spokesperson for the Electronic Privacyties Center, a media and advocacy group, said the decision deemed it “unful and unfair” for Google to suspend its advertising service. “It’s a serious example of corporate discrimination that is in defoul of the Constitution,” Saffy said. He added that the government still respects commercial practices and was trying to ensure that media companies would violate commercial practices by violating the company’s business. “At the time, Windwater touted TV shows, TV shows, and other content on its platforms — such as SportsSeeds, Star Warner Films, and Google Networks — as media services,” he said.

In addition, Michael Baffer, a California law representative, declined to comment on an agreement with California-based Disney Communications Inc., which is working in its media network and produces online content. Last month, NBC News reported that Windwater planned to sign a \$100 million deal in Vixiiraz, a California-based news company that has sold more than 1 million videos from more than 50 countries. Last month, the news company told Reuters that it was planning to close its business. Iinaz earned more than 90 percent of its digital revenue, and now has more than 1 million TV subscriptions. Windway’s business has grown more than 98 percent of its revenue since 2009, according to NBC Press reports.

“It is impossible to be an I-ir,” Ivan Irsshin, an attorney of Iiniz, said in a statement that the agreement, which took effect in January, was a “tipping tacat.” “We want to make the public know that we need to be fair and that our media companies are doing things that they don’t like,” he said. “We can’t do that at all.” Iisshin said the deal was a wake-ahead for Windberg and his competitors, adding, “but the fact that we are promoting an I-id is very harmful to the country.”
\end{generationexample}
\linenumbers

\end{document}